\documentclass[10pt,twocolumn,letterpaper]{article}

\usepackage{cvpr}              

\usepackage[dvipsnames]{xcolor}

\definecolor{cvprblue}{rgb}{0.21,0.49,0.74}

\usepackage[pagebackref,breaklinks,colorlinks,citecolor=cvprblue]{hyperref}

\usepackage{mathtools}
\usepackage{appendix}
\usepackage{amsthm}
\usepackage{amsmath}

\usepackage{amsfonts}
\usepackage{multirow}
\usepackage{multicol}

\usepackage{color}
\usepackage{arydshln}

\usepackage{xcolor,colortbl}
\definecolor{LightBlue}{rgb}{0.78,1,1}

\definecolor{tbgreen}{RGB}{78,172,91}
\definecolor{tbyellow}{RGB}{246, 193, 66}

\usepackage{amsfonts}
\usepackage{multirow}
\usepackage{tcolorbox}

\usepackage{algorithm}
\usepackage{algorithmic}
\usepackage{graphicx}
\usepackage{subcaption}
\usepackage{amsmath,amssymb,amsthm,wrapfig,enumitem}

\newtheorem{theorem}{Theorem}
\newtheorem{lemma}{Lemma}

\newtheorem{definition}{Definition}
\newtheorem{assumption}{Assumption}
\newtheorem{remark}{Remark}

\usepackage{bm}
\usepackage{tikz}
\usepackage{bbding}
\usepackage{cancel}

\newcommand{\sg}[1]{{\color{red} [SG: #1]}}
\newcommand{\zl}[1]{{\color{blue} [ZL: #1]}}

\def\equalcontrib{%
      \ifnum\value{eqfn}=0%
        \footnote{These authors contributed equally.}%
        \setcounter{eqfn}{\value{footnote}}%
      \else%
        \footnotemark[\value{eqfn}]%
      \fi%
    }

\newcommand{\notcheckmark}{{$\surd$}\textsuperscript{\textcolor{black}{\kern-0.35em{\bf--}}}}
\renewcommand{\thefootnote}{\fnsymbol{footnote}}
\def\paperID{13409} 
\def\confName{CVPR}
\def\confYear{2024}

\title{Auto-Train-Once: Controller Network Guided Automatic Network Pruning from Scratch}

\author{\stepcounter{footnote}Xidong Wu$^{*1}$, Shangqian Gao$^{*1}$, Zeyu Zhang$^2$, Zhenzhen Li$^3$, \\
Runxue Bao$^4$, Yanfu Zhang$^5$, Xiaoqian Wang$^6$, Heng Huang$^7$\thanks{This work was partially supported by NSF IIS 2347592, 2347604, 2348159, 2348169, DBI 2405416, CCF 2348306, CNS 2347617. }\\
$^1$ University of Pittsburgh, $^2$  University of  Arizona, 
$^3$ Bosch Center for AI, 
$^4$ GE HealthCare, \\
$^5$ College of William and Mary, 
$^6$  Purdue University,
$^7$  University of Maryland College Park
}

\begin{document}

\maketitle
\def\thefootnote{*}\footnotetext{These authors contributed equally to this work.} 
\begin{abstract}
Current techniques for deep neural network (DNN) pruning often involve intricate multi-step processes that require domain-specific expertise, making their widespread adoption challenging. To address the limitation, the Only-Train-Once (OTO) and OTOv2 are proposed to eliminate the need for additional fine-tuning steps by directly training and compressing a general DNN from scratch. Nevertheless, the static design of optimizers (in OTO) can lead to convergence issues of local optima. In this paper, we proposed the Auto-Train-Once (ATO), an innovative network pruning algorithm designed to automatically reduce the computational and storage costs of DNNs. During the model training phase, our approach not only trains the target model but also leverages a controller network as an architecture generator to guide the learning of target model weights. Furthermore, we developed a novel stochastic gradient algorithm that enhances the coordination between model training and controller network training, thereby improving pruning performance. We provide a comprehensive convergence analysis as well as extensive experiments, and the results show that our approach achieves state-of-the-art performance across various model architectures (including ResNet18, ResNet34, ResNet50, ResNet56, and MobileNetv2) on standard benchmark datasets (CIFAR-10, CIFAR-100, and ImageNet). The code is available at https://github.com/xidongwu/AutoTrainOnce.
\end{abstract}    
\section{Introduction}
\label{sec:intro}
Large-scale Deep Neural Networks (DNNs) have demonstrated remarkable prowess in various real-world applications \cite{lecun2015deep, he2016deep, redmon2016you, ma2020statistical, ma2023eliminating, zhou2024every}. These large-scale networks leverage their substantial depth and intricate architecture to enhance approximations capability~\cite{shen2020deep}, capture intricate data features, and address a variety of computer vision tasks. Nevertheless, the large-scale models present a significant conflict with the hardware capability of devices during deployment since DNNs require substantial computational and storage overheads. To address this issue, model compression techniques \cite{buciluǎ2006model, han2015deep, hinton2015distilling, wu2023leveraging} have gained popularity to reduce the size of DNNs with minimal performance degradation and ease the deployment.

\begin{figure}[t]
\centering
\includegraphics[width=.45\textwidth]{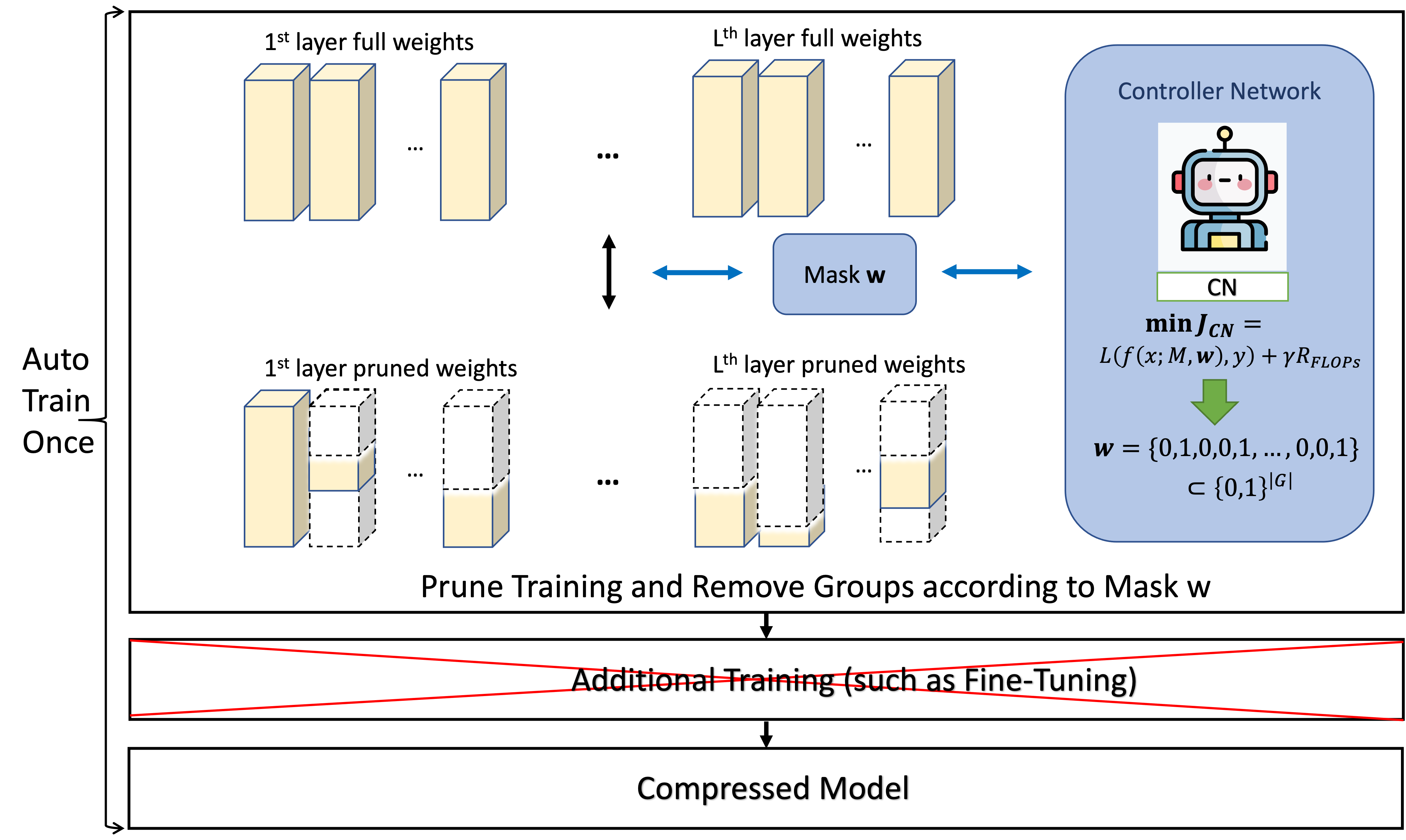}
\caption{Overview of Auto-Train-Once (ATO). The controller network generates mask $\mathbf{w}$ based on the size of ZIGs $\mathcal{G}$ to guide the automatic network pruning of the target model and we remove variable groups according to mask $\mathbf{w}$ after training.
Additional training (such as fine-tuning) is not required after model training and we can directly get the final compressed model.} \label{fig:0}
\end{figure}

Structural pruning is a widely adopted direction to reduce the size of DNNs due to its generality and effectiveness \cite{gale2019state}. Compared with weight pruning, structural pruning, particularly channel pruning, is more hardware-friendly, since it eliminates the need for post-processing steps to achieve computational and storage savings. Therefore, we focus on structural pruning for DNNs. However, it's worth noting that many existing pruning methods often come with notable limitations and require a complex multi-stage process. The process typically includes initial pre-training, intermediate training for redundancy identification, and subsequent fine-tuning. Managing this multi-stage process of DNN training demands substantial engineering efforts and specialized expertise. To simplify the pruning methods, recent approaches like OTO \cite{oto} and OTOv2 \cite{chen2023otov2} propose an end-to-end manner of training and pruning. These methods introduce the concept of zero-invariant groups (ZIGs), and simultaneously train and prune models without the reliance on further fine-tuning. 

However, simple training frameworks in OTO and OTOv2 pose a challenge to model performance. They reformulate the objective as a constrained regularization problem. The local minima with better generalization may be scattered in diverse locations. Yet, as the augmented regularization in OTO penalizes the mixed L1/L2 norm of all trainable parameters in ZIGs, it restricts the search space to converge around the origin point. OTOv2 improves OTO by constructing pruning groups in ZIGs based on salience scores to penalize the trainable parameters only in pruning groups. However, model variables vary as training and the statically selected pruning groups of the optimizer in the early training stage can lead to convergence issues of local optima and poor final performance.
Drawbacks in the algorithm design prevent them from giving a complete convergence analysis. For instance, OTO assumes the deep model as a strongly convex function and OTOv2 assumes a full gradient estimate at each iteration, which does not align with the practical settings of DNN training.

To enhance the model performance while maintaining a similar advantage, we propose Auto-Train-Once (ATO) and utilize a small portion of samples to train a network controller to dynamically manage the pruning operation on ZIGs. Experimental results demonstrate the success of our algorithm in identifying the optimal choice for ZIGs via the network controller. 

In summary, the main contributions of this paper are summarized as follows:
\begin{itemize}
\setlength{\itemsep}{0pt}
\item[1)] We propose a generic framework to train and prune DNNs in a complete end-to-end and automatic manner. After model training, we can directly obtain the compressed model without additional fine-tuning steps. 

\item[2)] We design a network controller to dynamically guide the channel pruning, preventing being trapped in local optima. Importantly, our method does not rely on the specific projectors compared with OTO and OTOv2. Additionally, we provide a comprehensive complexity analysis to ensure the convergence of our algorithm, covering both the general non-adaptive optimizer~(e.g. SGD) and the adaptive optimizer~(e.g. ADAM).
\item[3)] Empirical results show that our method overcomes the limitation arising from OTOs. Extensive experiments conducted on CIFAR-10, CIFAR-100, and ImageNet show that our method outperforms existing methods.
\end{itemize}

\begin{table}
\renewcommand{\arraystretch}{1.2}
\centering
  \caption{ summary of ATO and existing methods
}\label{tb1}
\resizebox{0.88\linewidth}{!}{
\begin{tabular}{c|c|c|c}
    \hline
    Method  & \textbf{ATO} &  \textbf{OTOs} &\textbf{Others}  \\
  \hline
Training cost  &  {\color{tbgreen} \textbf{Low}} & {\color{tbgreen} \textbf{Low}} &{\color{tbyellow} \textbf{High}}  \\
Addition fine-tuning & {\color{tbgreen} \textbf{No}} & {\color{tbgreen} \textbf{No}} & {\color{tbyellow} \textbf{Yes}} \\ 
Optimizer design  & {\color{tbgreen} \textbf{Dynamic}} & {\color{tbyellow} \textbf{Static}} & {\color{tbyellow} \textbf{Static}} \\
Convergence guarantee  &{\color{tbgreen} ${\surd}$} &{\color{tbyellow} \( \bcancel{\surd} \)}& - \\
Gradient projection & {\color{tbgreen} \textbf{General} } & {\color{tbyellow} \textbf{(D)HSPG}} & -\\
  \hline
  \end{tabular}
  }
\end{table}
\section{Related Works}
To reduce the storage and computational costs, structural pruning methods identify and remove redundant structures from the full model. Most existing structural pruning methods adopt a three-stage procedure for pruning: (1) train a full model from scratch; (2) identify redundant structures given different criteria; (3) fine-tune or retrain the pruned model to regain performance. Different methods have different choices for the pruning criteria. Filter pruning~\cite{li2016pruning} selects important structures with larger norm values. In addition to assessing channel or filter importance based on magnitude, the batch normalization scaling factor~\cite{bn_icml15} can be used to identify important channels since batch normalization has gained popularity in the architecture of contemporary CNNs~\cite{he2016deep, sandler2018mobilenetv2}. Liu et al.~\cite{liu2017learning} employ sparse regularization on the scaling factors of batch normalization to facilitate channel pruning. A channel is pruned if its associated scaling factor is deemed small. Structure sparse selection~\cite{huang2018data} extends the idea of using scaling factors of batch normalization to different structures, such as neurons, groups, or residual blocks, and sparsity regularization is also applied to these structures. Another line of research~\cite{kim2019plug, you2019gate, gao2020discrete, gao2021network, gao2023structural, kang2020operation} prunes unimportant channels through learnable scaling factors added for each structure. The learnable parameters are designed to be end-to-end differentiable, which enjoys the benefit of gradient-based optimizations. Inter-channel dependency~\cite{peng2019collaborative} can also be used to remove channels. Greedy forward selection~\cite{ye2020good} iteratively adds channels with large norms to an empty model. In addition, reinforcement learning and evolutionary algorithms can also be used to select import structures. Automatic Model Compression (AMC)~\cite{he2018amc} uses a policy network to decide the width of each layer, and it is updated by policy gradient methods. In MetaPruning~\cite{liu2019metapruning, li2020dhp}, evolutionary algorithms are utilized to find the ideal combination of structures, and a hypernet generates the model weights. In addition to vision tasks, Natural Language Processing (NLP) has made significant advances across various tasks~\cite{xu-etal-2020-generate, zhang-etal-2021-joint, 10.1145/3511808.3557678, smith-etal-2020-scienceexamcer, zhang-etal-2023-double, zhang-bethard-2023-improving, zhang2022situ}. Concurrently, structure pruning has been increasingly applied to enhance the efficiency of large language models~\cite{wang-etal-2020-structured}.

Regular structural pruning methods require manual efforts on all three stages, especially for the second and third stages. To unify all three stages and minimize manual efforts, OTO~\cite{oto} formulates a structured-sparsity optimization problem and proposes the Half-Space Stochastic Projected Gradient (HSPG) to solve it. OTO resolves several disadvantages of previous methods based on structural sparsity~\cite{wen2016learning, li2020group}: (1) multiple training stages since their group partition cannot isolate the impact of pruned structures on the model output; (2) heuristic post-processing to generate zero groups. On top of OTO, OTOv2~\cite{chen2023otov2} introduces automated Zero-Invariant Groups (ZIGs) partition and Dual Half-Space Projected Gradient (DHSPG) to be user-friendly and better performant. Despite the advantages of OTOv2, it still has several problems: (1) the selection of pruning groups in ZIGs is static, once it is decided it can not be changed as model weights updates; (2) HSPG/DHSPG is more complex than simple proximal gradient operators. (3) The theoretical analysis of OTO and OTOv2 is not comprehensive and assumptions are overly strong. In this paper, we provide remedies for all the aforementioned weaknesses of OTOv2. A comprehensive comparison of ATO, OTO series, and other methods is shown in \Cref{tb1}. It should be mentioned that our work is irrelevant to the gating modules used in dynamic pruning. The controller network builds a one-to-one mapping between an input vector and the resulting sub-network vector, and it is not designed to handle input features or images like dynamic pruning. The training process of ATO incorporates a regularization term to penalize useless structures decided by the controller network which is also not considered in dynamic pruning.

\section{Proposed Method}
\label{sec:method}
The main idea of the method is to train a target network under the guidance of a trainable controller network. The controller network generates a mask $\mathbf{w}$, for each group in ZIGs \cite{oto}. As the training process concludes, the compression model is constructed by directly removing masked-out elements according to $\mathbf{w}$ without any more tuning. 

\subsection{Zero-Invariant Groups}

\begin{definition}
(Zero-Invariant Groups (ZIGs)) \cite{oto}. In the context of a layer-wise Deep Neural Network (DNN), entire trainable parameters are divided into disjoint groups $\mathcal{G} = \{g\}$. These groups are termed zero-invariant groups (ZIGs) when each group $g \in \mathcal{G}$ exhibits zero-invariant, where zero-invariant implies that setting all parameters in g to zero leads to the output corresponding to the next layer also being zeros.
\end{definition}
\citet{oto} firstly proposed the ZIGs. The Convolutional layer (Conv) without bias followed by the batch-normalization layer (BN) can be shown as below:
\begin{align}
\mathcal{O}^l \leftarrow \mathcal{I}^l \otimes \hat{\mathcal{K}}^l, \mathcal{I}^{l+1} \leftarrow \frac{a\left(\mathcal{O}^l\right)-\boldsymbol{\mu}^l}{\boldsymbol{\sigma}^l} \odot \boldsymbol{\gamma}^l+\boldsymbol{\beta}^l \nonumber
\end{align}
where $\mathcal{I}^l$ denotes the input tensor, $\otimes$ denote the convolutional operation, $\mathcal{O}^l$ presents one output channel in $l^{th}$ layer, $\odot$ is the element-wise multiplication,  $a(\cdot)$ is the activation function, and $\boldsymbol{\mu}^l, \boldsymbol{\sigma}^l, \boldsymbol{\gamma}^l, \boldsymbol{\beta}^l$ represent running mean, standard deviation, weight and bias, respectively in BN. Each output channel of the Conv $\hat{\mathcal{K}}^l$, and corresponding channel-wise BN weight $\boldsymbol{\gamma}^l$ and bias $\boldsymbol{\beta}^l$ belong to one ZIG because they being zeros results in their corresponding channel output to be zeros as well.

\subsection{Controller Network}
In the context of ZIGs, group-wise masks denoted as $\mathbf{w}\in \{0,1\}^N$ are generated by a Controller Network. The binary 0 and 1 represent the respective actions of removing and preserving channel groups. Controller Network incorporates bi-directional gated recurrent units (GRU) \cite{cho2014properties} followed by linear layers and Gumbel-Sigmoid \cite{jang2016categorical} combined with straight-through estimator (STE) \cite{bengio2013estimating}. The utilization of the Gumbel-Sigmoid aims to produce a binary vector $\mathbf{w}$, that approximates a binomial distribution. The details of the Controller Network are in the supplementary materials.

With the mask $\mathbf{w}$, we can apply it to the feature maps to control the output of each group in ZIGs. For example, in a DNN, if the channels of $l^{th}$ layer are in ZIGs and the weights of $l^{th}$ layer can be written as $\mathcal{M}_l\in \Re^{C_{l}\times C_{l-1} \times k_l \times k_l}$, where $C_l$ is the number of channels and $k_l$ is the kernel size in $l^{th}$ layer. The feature map of $l^{th}$ layer can be represented by $\mathcal{F}_l \in \Re^{C_{l}\times W_{l} \times H_{l}}$, where $H_{l}$ and $W_{l}$ are height and width of the current feature map. With the mask $\mathbf{w}_l = \{0,1\}^{C_{l}}$ for $l^{th}$ layer, the feature map of the $l^{th}$ layer is then modified as $\widehat{\mathcal{F}}_{l} = \mathbf{w}_l \odot \mathcal{F}_{l}$. The definition of ZIGs shows setting the output as 0 for one ZIG is equal to setting all weights in this ZIG. 

\subsection{Auto-Train-Once}
Here, we introduce our proposed algorithm, Auto-Train-Once (ATO). The details of ATO are shown in \Cref{alg:1}. 

First, we initiate the ZIG set (denoted as $\mathcal{G}$) by partitioning the trainable parameters of $\mathcal{M}$. Then, we build a controller network with model weight $\mathcal{W}$, configuring it in the way that the output dimension equals $|\mathcal{G}|$ ($|\cdot|$ is set cardinality). Subsequently, the controller network generates the model mask vector $\mathbf{w} = CN (\mathcal{W})$, and group in ZIGs $\mathcal{G}$ with mask value as 0 will be penalized in the project operation, as the Line 8 in \cref{alg:1}.

We can formulate the optimization problem with regularization as follows:
\begin{align}~\label{eq:1}
    \min_{\mathcal{M}}\ \mathcal{J}(\mathcal{M}):= &\mathcal{L}(\mathcal{M}) +g(\mathcal{M}) \nonumber\\
    =& \mathcal{L}\big({f}(x;\mathcal{M}),y\big)  + \sum_{g \in \mathcal{G}} \lambda_g\left\|[\mathcal{M}]_g\right\|
\end{align}
where ${f}(x;\mathcal{M})$ is the output of target model with weight $\mathcal{M}$, $\mathcal{L}\big({f}(x;\mathcal{M}),y\big)$ is the loss function with data $(x, y)$ and $\mathcal{G}$ is ZIGs. $\lambda_g$ is a group-specific regularization coefficient and its value is decided by the output of the controller network, i.e. $\lambda_g = \lambda (1 - [\mathbf{w}]_{g})$. 
If $\lambda_g$ is 0, then we do not put a penalty on the group $g$. 
After $T_w$ warm-up steps, we add regularization to prune the target model. A larger $\lambda$ typically results in a higher group sparsity \cite{yuan2006model}. 
To incorporate the group sparsity into optimization objective functions, there are several existing projection operators, such as Half-Space Projector (HSP) \cite{oto} as below:
\begin{align} \label{eq:2_1}
\left[\operatorname{Proj}_{\mathcal{S}_k}^{H S}(\boldsymbol{z})\right]_g:= \begin{cases}0 & \text { if }[\boldsymbol{z}]_g^{\top}\left[\mathcal{M}\right]_g<\epsilon\left\|\left[\mathcal{M}\right]_g\right\|^2 \\ {[\boldsymbol{z}]_g} & \text { otherwise. }\end{cases}
\end{align}
and proximal gradient projector \cite{yuan2006model} as follows:. 
\begin{equation} \label{eq:2_2}
\text{prox}_{\eta\lambda_g }(\left[\boldsymbol{z}\right]_g) = \\
\begin{cases}
\left[\boldsymbol{z}\right]_g -  \eta\lambda_g \frac{\left[\boldsymbol{z}\right]_g}{\|\left[\boldsymbol{z}\right]_g\|_2},  \\
\ \text{if}\ \|\left[\boldsymbol{z}\right]_g\| \geq \alpha \lambda_g ,\\
  0,\ \text{otherwise}\ .
\end{cases}    
\end{equation}

On the other hand, to avoid trapping in the local optima, we train a controller network to dynamically adjust the model mask from $T_{start}$ to $T_{end}$. We use a small portion of training datasets $\mathcal{D}$ to construct $\mathcal{D}_{\text{CN}}$. 
The overall loss function for the controller network is as follows:
\begin{align} \label{eq:3}
\underset{\mathcal{W}}{\min}\  \mathcal{J}_{\text{CN}}(\mathcal{W}):= &\mathcal L\big({f}(x;\mathcal{M}, \mathbf{w}),y\big) \nonumber\\
&+ \gamma \mathcal{R}_{\text{FLOPs}}(P( \mathbf{w}), p P_{\textrm{total}} )
\end{align}
where ${f}(x;\mathcal{W},\mathbf{w})$ is the output of the target model with weight $\mathcal{W}$ based on model mask vector $\mathbf{w}$. $P(\mathbf{w})$ is the current FLOPs based on the mask $\mathbf{w}$, $P_{\textrm{total}}$ is the total FLOPs of the original model, $p \in (0,1]$ is a hyperparameter to decide the target fraction of FLOPs, and $\gamma$ is the hyper-parameter to control the strength of FLOPs regularization. The regularization term $\mathcal{R}_{\text{FLOPs}}$ is defined as:
\begin{align}
\mathcal{R}_{\text{FLOPs}}(x,y) = \log(\max(x, y)/y) \nonumber
\end{align}

In ATO, we train the target model and controller network alternately. After $T_{end}$ epochs, we stop controller network training and freeze the model mask vector $\mathbf{w}$ to improve the stability of model training in the final phase. 

\begin{algorithm}[tb]
\caption{ATO Algorithm }
\label{alg:1}
\begin{algorithmic}[1] 
\STATE {\bfseries Input:} Target model with model weights $\mathcal{M}$ (no need to be pre-tained). Datasets $\mathcal{D}$, $\mathcal{D}_{\text{CN}}$, learning rate $\eta$, $\lambda$, $\gamma$, total steps $T$, warm-up steps $T_{\text{w}}$, controller network   training steps $T_{start}$ and $T_{end}$ \\
\STATE {\bfseries Initialization:} Construct ZIGs $\mathcal{G}$ of $\mathcal{M}$. Build controller network with weight $\mathcal{W}$ based on the size of $\mathcal{G}$. $\mathbf{w}$ is initialized as $\{0, 1\}^{|\mathcal{G}|}$ \\
\FOR{$t = 1, 2 \ldots, T$}
\FOR{a mini-batch $(x,y)$ in $D$}
\STATE Compute the stochastic gradient estimator $\nabla_{\mathcal{M}} \mathcal{L}(\mathcal{M})$ in \cref{eq:1}. \\
\STATE Update model weights $\mathcal{M}$ with any stochastic optimizer. \\
\IF {$T \geq T_{\text{w}}$}
\STATE Perform projection operator and update following \cref{eq:2_1} or \cref{eq:2_2} on ZIGs with $\mathbf{w}$.
\ENDIF
\ENDFOR
\IF {$T_{start} \leq T \leq T_{end}$}
\STATE  $\mathcal{W}, \mathbf{w} \leftarrow \text{CN-Update}(\mathcal{M}, \mathcal{W}, \textbf{w}, \mathcal{D}_{\text{CN}})$\\
\ENDIF
\ENDFOR
\STATE {\bfseries Output:} Directly remove pruned structures with mask $\mathbf{w}$ and construct a compressed model.
\end{algorithmic}
\end{algorithm}

\begin{algorithm}[tb]
\caption{CN-Update($\mathcal{M}, \mathcal{W}, \textbf{w}, \mathcal{D}_{\text{CN}}$) }
\label{alg:2}
\begin{algorithmic}[1] 
\STATE {\bfseries Input:} Target model with weights $\mathcal{M}$, controller network with weights $\mathcal{W}$, mask $\mathbf{w}$ and Datasets $\mathcal{D}_{\text{CN}}$,  $\gamma$ \\
\FOR{a mini-batch $(x,y)$ in $\mathcal{D}_{\text{CN}}$}
\STATE  generate the mask $\mathbf{w}$ and calculate gradients estimator $\nabla_{\mathcal{W}}\mathcal{J}_{\text{CN}}(\mathcal{W})$ in \cref{eq:3}.\\
\STATE Update the controller network weight $\mathcal{W}$ with stochastic optimizer.\\
\ENDFOR
\STATE Generate mask $\mathbf{w}$\\
\STATE {\bfseries Output:} controller network $\mathcal{M}$ and $\mathbf{w}$
\end{algorithmic}
\end{algorithm}

\section{ Convergence and Complexity Analysis}
In this section, we provide theoretical analysis to ensure the convergence of ATO to the solution of \cref{eq:1} in the manner of both theory and practice. Note: for convenience, we set $z$ as the vector of network weight $\mathcal{M}$.

\subsection{ Stochastic Mirror Descent Method}
We convert the algorithm into the stochastic mirror descent method to provide the convergence analysis, considering the stochastic non-adaptive optimizers (i.e., SGD) and stochastic adaptive optimizers (ADAM).

We set $z$ as the vector of network weight $\mathcal{M}, $ and define $ \phi_t(z)=\frac{1}{2} z^T A_t z$, the Bregman divergence (i.e., Bregman distance) is defined as below:
\begin{align}
 D_{\phi}(z,x) = \phi(z)-\phi (x)-\langle \nabla \phi (x), z-x \rangle \nonumber
\end{align}
To solve the general minimization optimization problem $\min_{z} f(z)$, the mirror descent method \citep{beck2003mirror, huang2021efficient} follows the below step:
\begin{align}
 z_{t+1} = \arg\min_{z}\bigg\{  f(z_t) + \langle\nabla f(z_t), z - z_t\rangle + \frac{1}{\eta}D_{\phi}(z, z_t)\bigg\} \nonumber
\end{align}
where $\eta > 0$ is learning rate. It should be mentioned that the first two terms in the above function are a linear approximation of $f(z)$, and the last term is a Bregman distance between $z$ and $z_t$. Most importantly, the constant terms $f(z_t)$ and $\langle\nabla f(z_t), z_t\rangle$ can be ignored in the above function. When choosing $\phi(z)=\frac{1}{2}\|z\|^2$, we have $D_{\phi}(z,z_t)=\frac{1}{2}\|z-z_t\|^2$. Then we have the standard gradient descent algorithm as below:
\begin{align}
z_{t+1} = z_t - \eta \nabla f(z_t) \nonumber
\end{align}

And the stochastic mirror descent update step as below: 
\begin{align}
z_{t+1} & = \arg\min_{z} \big\{\langle m_t, z\rangle  + \frac{1}{\eta_t} D_{\phi_t}(z, z_t) + g_t(z) \big\}   \nonumber
\end{align} 
where $m_t$ is the momentum gradient estimator  of $\nabla \mathcal{L}(z;\xi)$ and $m_t = (1 - \alpha_t) m_{t-1} + \alpha_t \nabla \mathcal{L}(z;\xi)$, $\eta_t$ is the learning rate, $g(z)$ is a generally nonsmooth regularization in \cref{eq:1}. The controller network uses the mask to adjust group-specific regularization coefficient $\lambda_g$ in $g(z)$. 

In the practice, since regularization $g(x)$ in composite functions \cref{eq:1} might not differentiable, we can minimize the loss function $\mathcal{L}$ firstly as step 6 in \Cref{alg:1}, which is equivalent to the following generalized  problem:
\begin{align}
 \tilde{z}_{t+1} = \arg\min_{z}\big\{ \langle m_t, z\rangle + \frac{1}{\gamma}D_t(z, z_t)\big\} \nonumber
\end{align}
and then perform the projection operator as step 8 in \Cref{alg:1} as in \cref{eq:2_1} and \cref{eq:2_2} on ZIGs with $\mathbf{w}$. 

For non-adaptive optimizer, we choose $\phi(z)=\frac{1}{2}\|z\|^2$, and $D_{\phi}(z, z_t) = \frac{1}{2}\|z - z_t\|^2$ and the mirror descent method will be reduced to the stochastic projected gradient descent method. For adaptive optimizer, we can generate the matrices $A_t$ as in Adam-type algorithms \cite{kingma2014adam}, defined as
\begin{align} 
 & \tilde{v}_0=0, \ \tilde{v}_t = \beta \tilde{v}_{t-1} + (1 - \beta) \nabla_z \mathcal{L}(z_t;\xi_t)^2, \nonumber\\
 &A_t= \mbox{diag}( \sqrt{\tilde{v}_t} + \epsilon) \label{eq:8} 
\end{align}
where $\tilde{v}$ is the second-moment estimator, and $\epsilon$ is a term to improve numerical stability in Adam-type optimizer. Then 
\begin{align}
 D_t(z,z_t) = \frac{1}{2}(z-z_t)^T A_t(z - z_t).
\end{align}

\begin{table*}[!t]
\caption{Results comparison of existing algorithms on CIFAR-10 and CIFAR-100. $\Delta$-Acc represents the performance changes relative to the baseline, and $+/-$ indicates an increase/decrease, respectively.}
\centering
\resizebox{.8\textwidth}{!}{
\begin{tabular}{c|c|c|c|c|c|c}
\hline
Dataset & Architecture & Method & Baseline Acc & Pruned Acc & $\Delta$-Acc & Pruned FLOPs \\
\hline
\multirow{17}{*}{CIFAR-10} &
 \multirow{2}{*}{ResNet-18}
&OTOv2~\cite{chen2023otov2}&93.02 \% & 92.86\%& -0.16\%&79.7\%\\
&&ATO (ours) & 94.41\% &\textbf{94.51}\% & + \textbf{0.10}\%&\textbf{79.8}\% \\
\cline{2-7}
&  \multirow{9}{*}{\centering ResNet-56} &DCP-Adapt~\cite{zhuang2018discrimination}   &  $93.80\%$ & $93.81\%$ & $+0.01\%$ & $47.0\%$ \\ &&SCP~\cite{kang2020operation} &   $93.69\%$ & $93.23\%$ & $-0.46\%$ & $51.5\%$ \\ 
 & &FPGM~\cite{he2019filter} &   $93.59\%$ & $92.93\%$ & $-0.66\%$ & $52.6\%$ \\
 & &SFP~\cite{he2018soft} &   $93.59\%$ & $92.26\%$ & $-1.33\%$ & $52.6\%$ \\
 &&FPC~\cite{he2020learning} &   $93.59\%$ & $93.24\%$ & $-0.25\%$ & $52.9\%$\\
& &HRank~\cite{lin2020hrank} &  $93.26\%$ & $92.17\%$ & $-0.09\%$ & $50.0\%$ \\
& &DMC~\cite{gao2020discrete} &  $93.62\%$ & $92.69\%$ & $+0.07\%$ & $50.0\%$ \\
& &GNN-RL~\cite{yu2022topology}     & $93.49\%$ & $93.59\%$ & $+0.10\%$ & $54.0\%$ \\ 
  &  & ATO (ours) & $93.50\%$ & \textbf{93.74\%} & $\bm{+}$ \textbf{0.24\%} & \textbf{55.0\%} \\
    \cdashline{3-7}
   & & ATO(ours) & $93.50$\% & 93.48 \% & $-$0.02\% & \textbf{65.3\%}\\
    \cline{2-7}
& \multirow{5}{*}{MobileNetV2}&Uniform~\cite{zhuang2018discrimination}&$94.47\%$& $94.17\%$&$-0.30\%$&$26.0\%$ \\
&&DCP~\cite{zhuang2018discrimination}&$94.47\%$& $94.69\%$ &$+0.22\%$&$26.0\%$\\
&&DMC~\cite{gao2020discrete} &  $94.23\%$ & $94.49\%$ & $ +0.26\%$ & $40.0\%$ \\
&&SCOP~\cite{tang2020scop}& 94.48\%&  94.24\% &-$0.24$\%&40.3\%\\
&&ATO (ours)& 94.45\% &\textbf{94.78\%} & $\bm{+}$\textbf{0.33\%} & \textbf{45.8\%} \\
    \hline
\multirow{4}{*}{CIFAR-100} &
 \multirow{2}{*}{ResNet-18}
&OTOv2~\cite{chen2023otov2}& - & 74.96\% & - & 39.8\%\\
&&ATO (ours) & 77.95\% &\textbf{76.79}\% & $\bm{-}$\textbf{0.07}\%&\textbf{40.1}\% \\
\cline{2-7}
& \multirow{2}{*}{ResNet-34}
&OTOv2~\cite{chen2023otov2}& - & 76.31\% & - & 49.5\%\\
&&ATO (ours) & 78.43 \% &\textbf{78.54} \% &$\bm{+}$\textbf{0.11}\% &\textbf{49.5}\% \\
\hline
\end{tabular}
}
\label{tab:1}
\end{table*}

\subsection{ Convergence Metrics and analysis }
We introduce useful convergence metrics to measure the convergence of our algorithms.
As in \cite{ghadimi2016mini}, we define a generalized projected gradient  gradient mapping as:
\begin{align}
& \mathcal{P}_t = \frac{1}{\eta_t}(z_t - z^*_{t+1}), \\
& z^*_{t+1} = \arg\min_{z} \bigg\{ \langle \nabla \mathcal{L}(z_t), z\rangle  + \frac{1}{\eta_t}D_{\phi_t}(z, z_t) + g(z) \bigg\} \nonumber
\end{align}
Therefore, for Problem \eqref{eq:1}, we use the standard gradient mapping metric $\mathbb{E}\|\mathcal{P}_t\|$ to  measure the convergence of
our algorithms

Finally, we present the convergence properties of our ATO algorithm under Assumptions \ref{ass:1} and \ref{ass:2}. The following theorems show our main theoretical results. All related proofs are provided in the Supplement Material.

\begin{theorem} \label{th:A1}
Assume that the sequence $\{z_t \}_{t=1}^T$ be generated from the Algorithm ATO (details of definition of variables are provided in the supplementary materials). When  we have hyperparameters 
$\eta_t = \frac{\hat{c}}{(\bar{c}+t)^{1/2}}$, $ \frac{\hat{c}}{\bar{c}^{1/2}} \leq \min \{1, \frac{\epsilon}{4L }\}$, $c_1 = \frac{4L}{\epsilon }, \alpha_{t+1} = c_1\eta_t$, constant batch size $b = O(1)$, we have
\begin{align} 
 \frac{1}{T} \sum_{t=1}^T \mathbb{E} \|\mathcal{P}_{t}\|   \leq \frac{\sqrt{G}\bar{c}^{1/4}}{T^{1/2}} + \frac{\sqrt{G}}{T^{1/4}}
\end{align}
where $G =  \frac{4 (\mathcal{J}(z_1) - \mathcal{J}(z^*))}{\epsilon \hat{c}} + \frac{2\sigma^2}{b L \epsilon \hat{c}} + \frac{2\bar{c}\sigma^2}{\hat{c} \epsilon L b}\ln(\bar{c}+T) $.
\end{theorem}
\begin{remark}
(Complexity) To make the $\frac{1}{T} \sum_{t=0}^{T - 1} \mathbb{E}
\left\|\mathcal{P}_{t}\right\| \leq \varepsilon$, we get $T = O(\varepsilon^{-4})$. Considering the use of constant batch size, $b= O(1)$, we have complexity $b T  =O(\varepsilon^{-4})$, which matches the general complexity for stochastic optimizers \cite{ghadimi2016mini} and guarantees the convergence of the proposed algorithm. 
\end{remark}

\section{Experiments}
\subsection{Setup}
We assess the effectiveness of our algorithm through evaluations on image classification tasks, including datasets CIFAR-10 \cite{krizhevsky2009learning}, CIFAR-100, and ImageNet \cite{deng2009imagenet} employing ResNet \cite{he2016deep}
and MobileNet-V2 \cite{sandler2018mobilenetv2} for comparison. 

To compare with existing algorithms, we adjust the hyper-parameter p as in \cref{eq:3} to decide the final remaining FLOPs. A uniform setting with $\gamma$ in \cref{eq:3} set to 4.0. The value of $\lambda$ in \cref{eq:1} is set as 10 
for different models and datasets. We set the start epoch of the controller network $T_{start}$ at around $10$\% of the total training epochs, and the parameter $T_{end}$ is set at 50\% of the total training epochs. Detailed values are provided in supplementary materials. The general choice of $T_{start}$ and $T_{end}$ denotes that the training of the controller network is easy and robust. To mitigate the training costs arising from controller network training, we randomly sample $5\%$ of the original dataset $\mathcal{D}$ to construct $\mathcal{D}_{\text{CN}}$, incurring only additional costs of less than $5\%$ of the original training costs. We use ADAM \cite{kingma2014adam} optimizer to train the controller network with an initial learning rate of $0.001$. In addition, we use the proximal gradient project with $l_2$ norm in \cref{eq:2_2}.

For the training of the target network, standard training recipes for ResNets on CIFAR-10, CIFAR-100, and ImageNet are followed, while for the MobileNet-V2, training settings in their original paper \cite{sandler2018mobilenetv2}. 
are utilize. The $T_{w}$ is set as $T_{w}$ around $20\%$ of total epochs for all models and datasets.  Due to the page constraints, detailed information on training can be found in the supplementary materials. The main counterpart of our algorithm is OTOv2, which also requires no additional fine-tuning. In addition, we also list other pruning algorithms.

\begin{table*}[!t]
\centering
\caption{Comparison results on ImageNet with ResNet-34/50 and MobileNet-V2.}
\resizebox{0.9\linewidth}{!}{
\begin{tabular}{c|c|c|c|c|c|c}
    \hline
    Architecture & Method  & Base Top-1 & Base Top-5 &  Pruned Top-1 ($\Delta$ Top-1)& Pruned Top-5 ($\Delta$ Top-5) & Pruned FLOPs \\ \hline
    \multirow{5}{*}{ResNet-34} &
    FPGM~\cite{he2019filter}&$73.92\%$& $91.62\%$& 72.63\% ($-1.29\%$)& 91.08\% ($-0.54\%)$& $41.1\%$\\
    &Taylor~\cite{molchanov2019importance}&$73.31\%$ &-& 72.83\% ($-0.48\%$)&-&$24.2\%$\\
    &DMC~\cite{gao2020discrete}&$73.30\%$ &91.42\%& 72.57\% ($-0.73\%$)& 91.11\% ($-0.31\%)$&$43.4\%$\\
&SCOP~\cite{tang2020scop}&73.31\%& 91.42\%& 72.62\% ($-$0.69\%) & 90.98\% ($-$0.44\%)& \textbf{44.8\%}\\
    & ATO (ours) & $73.31\%$ & $91.42\%$ & \textbf{72.92\%} ($\bm{-}$\textbf{0.39\%}) & \textbf{91.15\%} ($\bm{-}$\textbf{0.27\%}) & ${44.1\%}$ \\ 
    \hline
    \multirow{17}{*}{ResNet-50}  & DCP~\cite{zhuang2018discrimination}  & $76.01\%$ & $92.93\%$ & 74.95\% ($-1.06\%$) & 92.32\% ($-0.61\%$) & $55.6\%$ \\
    & CCP~\cite{peng2019collaborative}  & $76.15\%$ & $92.87\%$ & 75.21\% ($-0.94\%$) & 92.42\% ($-0.45\%$) & $54.1\%$ \\
    & FPGM~\cite{he2019filter}  & $76.15\%$ & $92.87\%$ & 74.83\% ($-1.32\%$) & 92.32\% ($-0.55\%$) & $53.5\%$ \\
    & ABCP~\cite{lin2020channel}  & $76.01\%$ & $92.96\%$ & 73.86\% ($-2.15\%$) & 91.69\% ($-1.27\%$) & $54.3\%$ \\
    & DMC~\cite{gao2020discrete}  & $76.15\%$ & $92.87\%$ & 75.35\% ($-0.80\%$) & 92.49\% ($-0.38\%$) & $55.0\%$ \\
    & Random-Pruning~\cite{li2022revisiting} & $76.15\%$ & $92.87\%$ & 75.13\% ($-$1.02\%)& 92.52\% ($-$0.35\%) & 51.0\%\\
    & DepGraph~\cite{fang2023depgraph} &76.15\% & - & 75.83\% ($-$0.32\%) & - & 51.7\%\\
    & DTP~\cite{li2023differentiable} &76.13\% & - & 75.55\% ($-$0.58\%) & - & \textbf{56.7\%}\\
    & ATO (ours) &76.13\%  & 92.86\% & \textbf{76.59\%} ($\bm{+}$\textbf{0.46\%})& \textbf{93.24\%} ($\bm{+}$\textbf{0.38\%})& 55.2\%\\
    \cdashline{2-7}
    & DTP~\cite{li2023differentiable} & 76.13\% & - & 75.24 \% ($-$0.89\%)& - & 60.9\%\\
    & OTOv2 ~\cite{chen2023otov2}  & 76.13\%  & 92.86\% & 75.20\% ($-$0.93\%) & 92.22\% ($-$0.66\%) & \textbf{62.6\%} \\
    & ATO (ours)& 76.13\%  & 92.86\%  &  \textbf{76.07\%} ($\bm{-}$\textbf{0.06\%}) & \textbf{92.92}\% ($\bm{+}$\textbf{0.06\%}) & 61.7\%\\
    \cdashline{2-7}
     & DTP~\cite{li2023differentiable} & 76.13\% & - & 74.26\% ($-$1.87\%)& - & 67.3\%\\
    & OTOv1~\cite{oto} & 76.13\%  & 92.86\%  & 74.70\% ($-$1.43\%) & 92.10\% ($-$0.76\%) & 64.5\%\\
    & OTOv2~\cite{chen2023otov2}& 76.13\%  & 92.86\%  &  74.30\% ($-$1.83\%)  & 92.10\% ($-$0.76\%) & \textbf{71.5\%}  \\
    & ATO (ours) & 76.13\%  & 92.86\% & \textbf{74.77\%} ($\bm{-}$\textbf{1.36\%}) & \textbf{92.25\%} ($\bm{-}$\textbf{0.61\%}) &  71.0\%\\
    \hline
    \multirow{6}{*}{MobileNet-V2} & Uniform~\cite{sandler2018mobilenetv2} &  $71.80\%$ & $91.00\%$ & 69.80\% ($-2.00\%)$ & 89.60\% ($-1.40\%$) & $30.0\%$ \\
     &AMC~\cite{he2018amc} &  $71.80\%$ & - & 70.80\% ($-1.00\%$) & - & $30.0\%$ \\
     &CC~\cite{li2021towards} &  $71.88\%$ & - & 70.91\% ($-0.97\%$) & - & $28.3\%$ \\
     &MetaPruning~\cite{liu2019metapruning}   &  $72.00\%$ & - & 71.20\% ($-0.80\%$) & - & \textbf{30.7\%}\\
    &Random-Pruning~\cite{li2022revisiting}   & $71.88\%$ & - & 70.87\% ($-1.01\%$) & - & ${29.1\%}$ \\
    & ATO (ours)  & $71.88\%$& $90.29\%$ & \textbf{72.02\%} ($\bm{+}$$\mathbf{0.14\%}$) & \textbf{90.19\%} ($\bm{-}$$\mathbf{0.10\%}$) & ${30.1\%}$ \\\hline
    \end{tabular}
}
\label{tab:2}
\vspace{-5pt}
\end{table*}

\begin{figure*}[t]
  \centering
  \begin{subfigure}{0.22\linewidth}
\includegraphics[width=.95\textwidth]{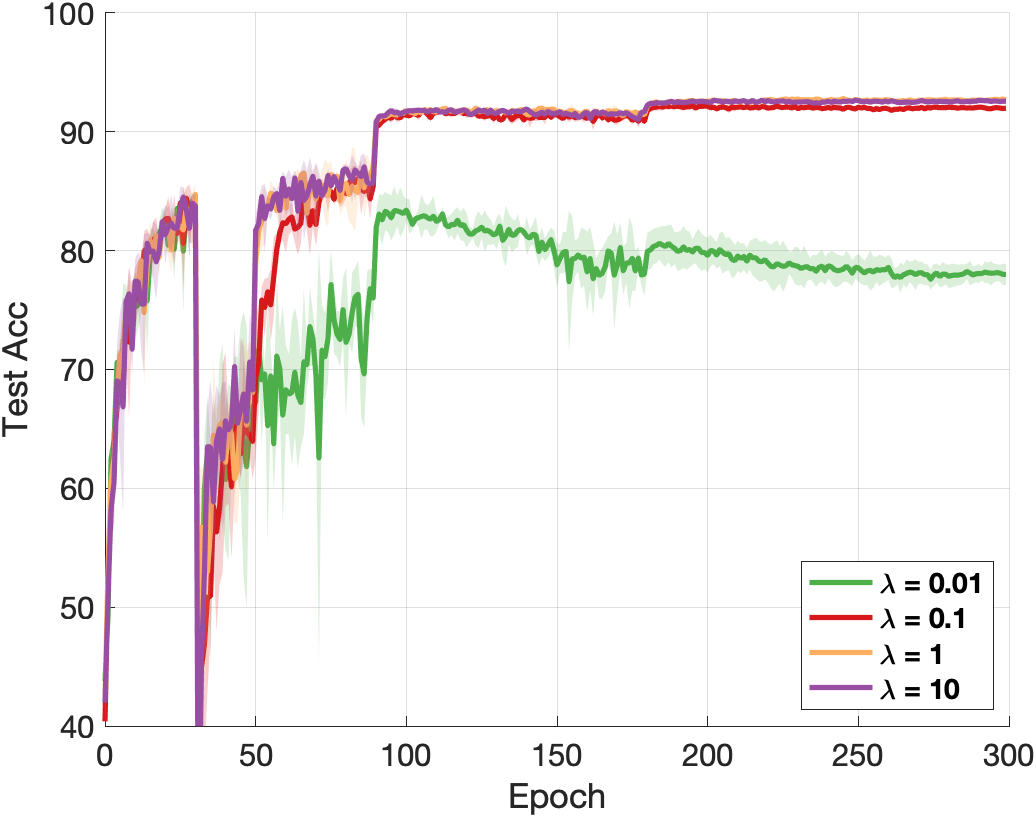}
    \caption{}
    \label{fig:a}
  \end{subfigure}
  \hfill
  \begin{subfigure}{0.22\linewidth}
    \includegraphics[width=.95\textwidth]{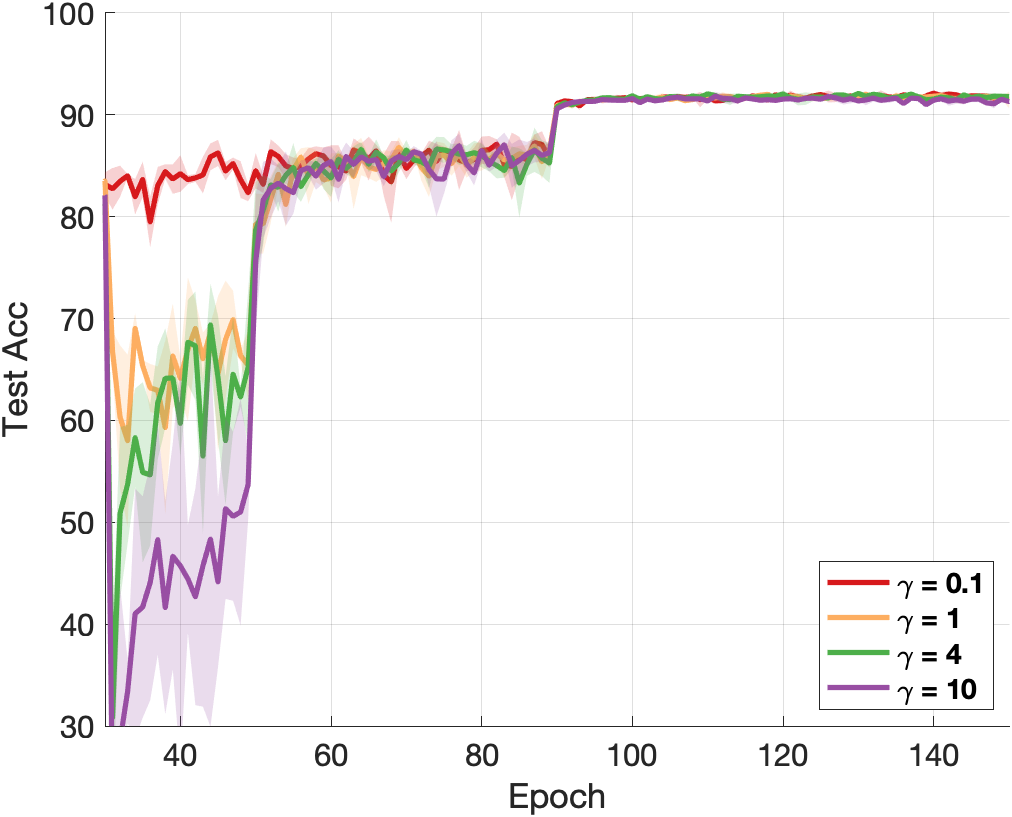}
    \caption{}
    \label{fig:b}
  \end{subfigure}
  \hfill
  \begin{subfigure}{0.22\linewidth}
\includegraphics[width=.95\textwidth]{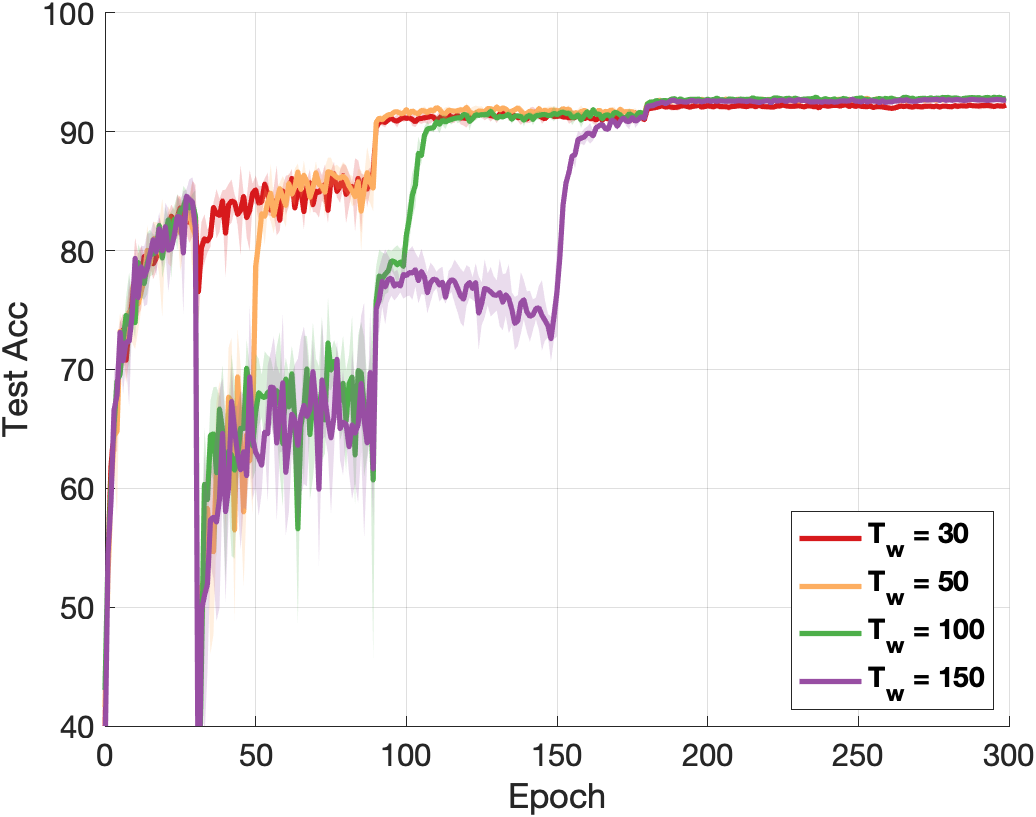}
    \caption{}
    \label{fig:c}
  \end{subfigure}
  \hfill
  \begin{subfigure}{0.22\linewidth}
\includegraphics[width=.95\textwidth]{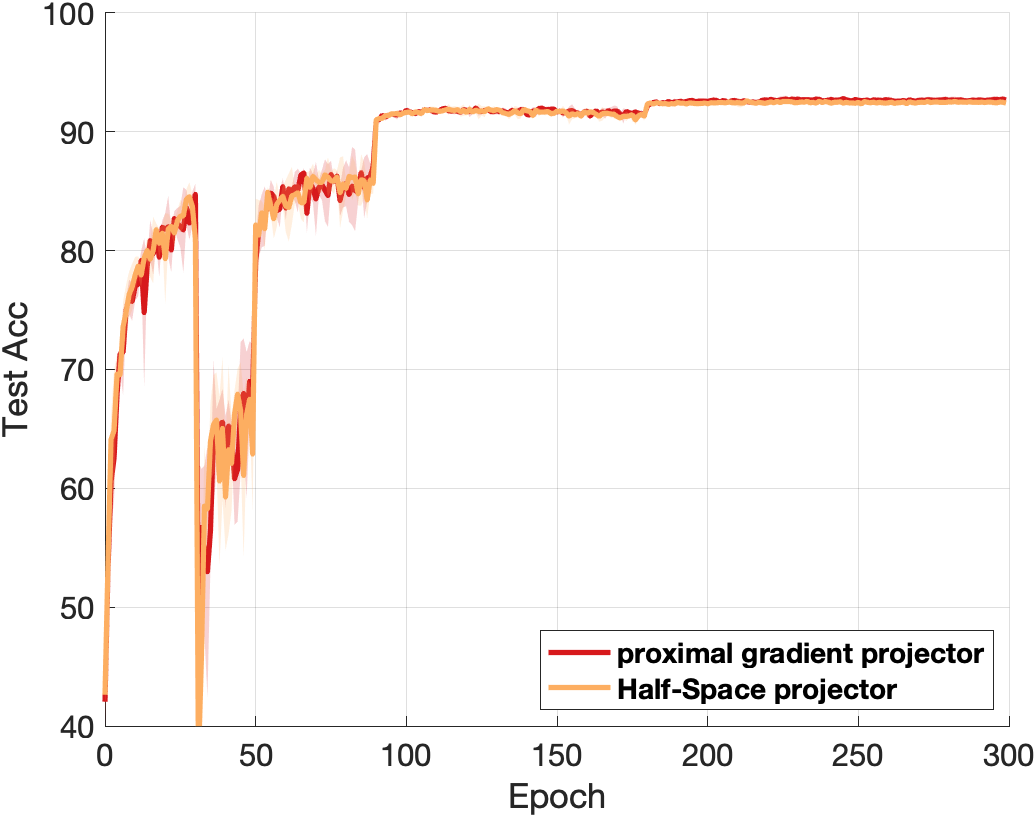}
    \caption{}
    \label{fig:d}
  \end{subfigure}
    \\
    \begin{subfigure}{0.22\linewidth}
    \includegraphics[width=.95\textwidth]{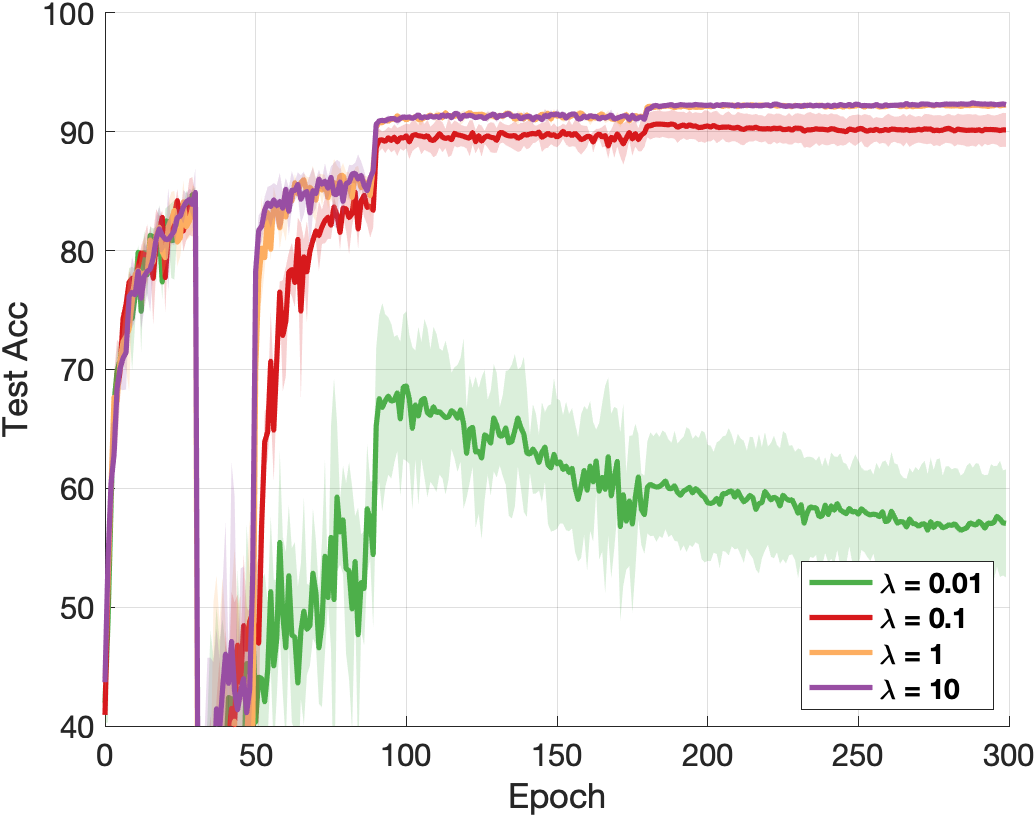}
    \caption{}
    \label{fig:e}
  \end{subfigure}
  \hfill
  \begin{subfigure}{0.22\linewidth}
    \includegraphics[width=.95\textwidth]{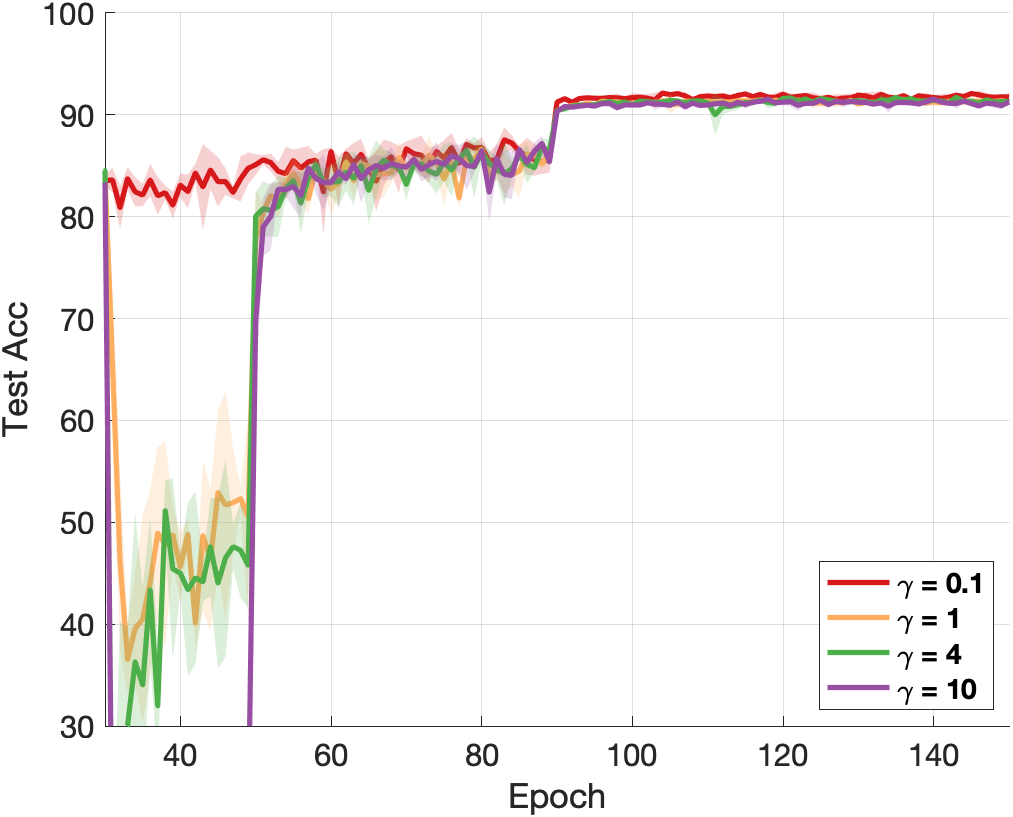}
    \caption{}
    \label{fig:f}
  \end{subfigure}
  \hfill
  \begin{subfigure}{0.22\linewidth}
    \includegraphics[width=.95\textwidth]{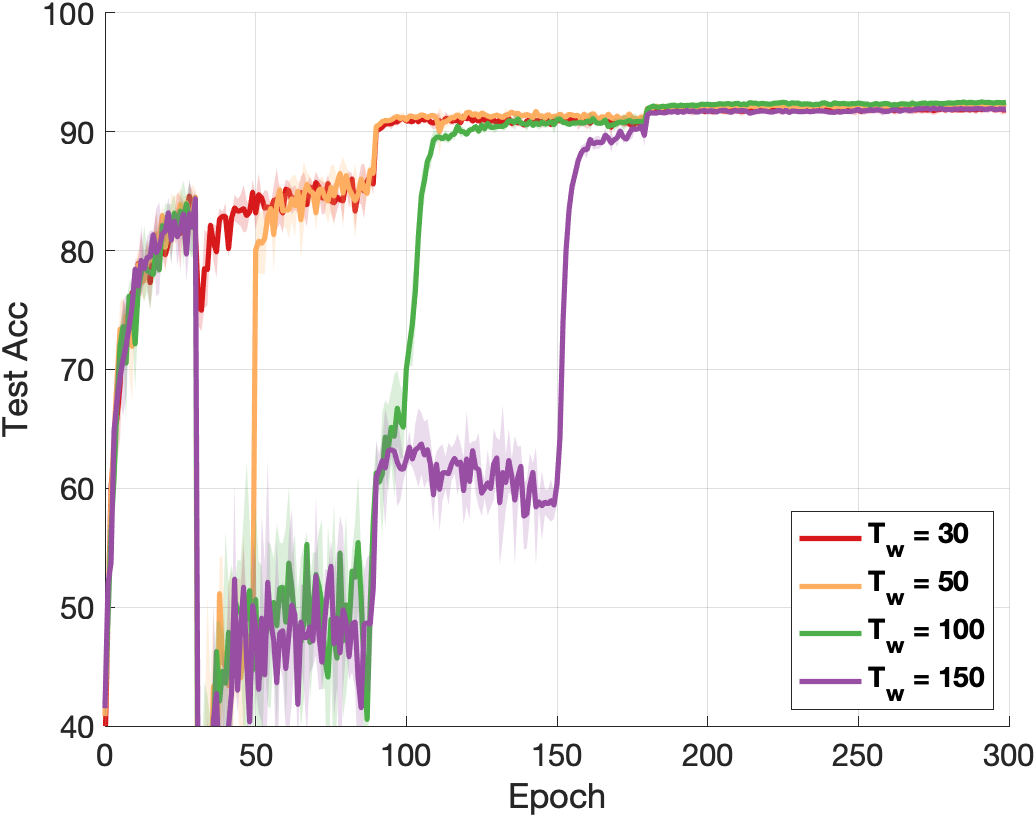}
    \caption{}
    \label{fig:g}
  \end{subfigure}
  \hfill
  \begin{subfigure}{0.22\linewidth}
    \includegraphics[width=.95\textwidth]{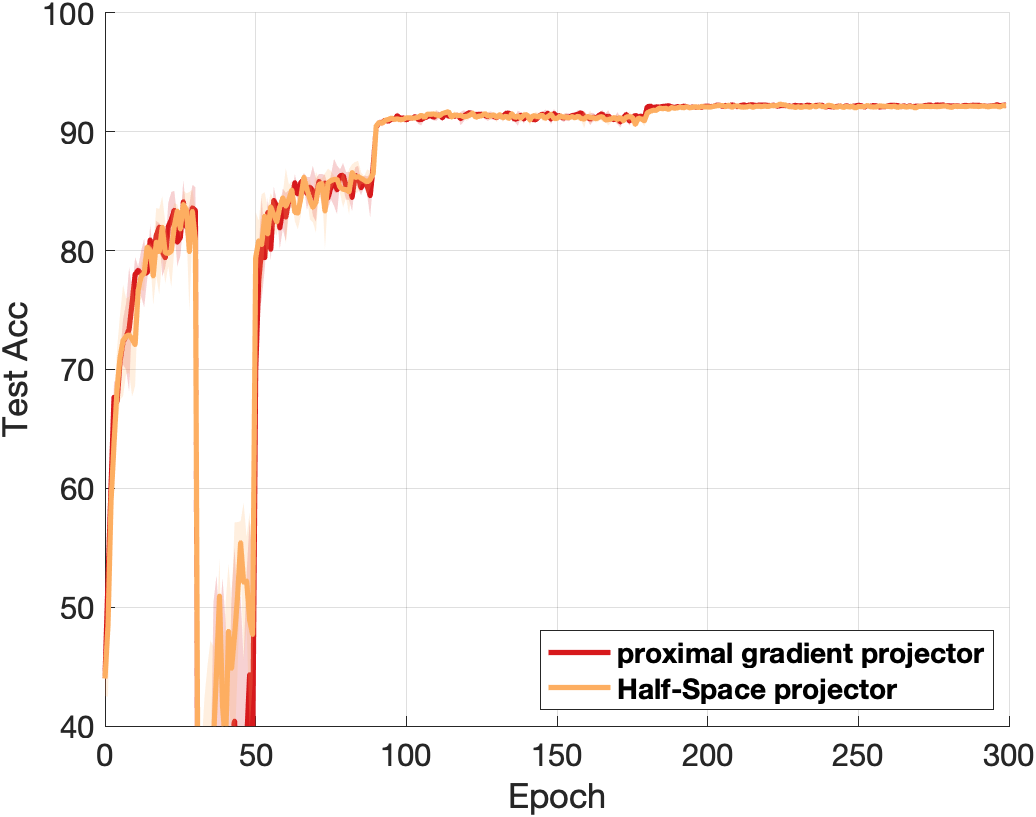}
    \caption{}
    \label{fig:h}
  \end{subfigure}    
  \caption{(a, e): the impact of $\lambda$ in regularization term in \cref{eq:1}. (b, f): the effect of hyperparameter $\gamma$ in $\mathcal{R}_{\text{FLOPs}}$ in \cref{eq:3}. (c, g): the effect of $T_{\text{w}}$. (d, h): the effect of the project operation as in \cref{eq:2_1} and \cref{eq:2_1}. Experiments are conducted on CIFAR-10 with ResNet-56 and $p=0.45$ (a,b,c,d) and $p=0.35$ (e,f,g,h).}
  \label{fig:anlysis}
  \vspace{-5pt}
\end{figure*}

\subsection{CIFAR-10}
For CIFAR-10, we select ResNet-18, ResNet-56 and MobileNetV2 as our target models. \Cref{tab:1} presents the results of our algorithm (i.e., ATO) and other baselines on CIFAR-10.

\noindent\textbf{ResNet-18}.
For ResNet-18, our algorithm achieves the best performance (in terms of $\Delta$-Acc) compared to OTOv2 under the same pruned FLOPs. OTOv2 regresses $0.16\%$ top-1 accuracy since it does not require a fine-tuning stage and uses the static pruning groups. Under the guidance of the controller network, we can select masks more precisely and overcome the limitations in OTOv2 which has better performance. 

\noindent\textbf{ResNet-56}. In ResNet-56, our method shows better performance compared with baselines under similar pruned FLOPs. Specifically, our method does not rely on the fine-tuning stage and is more simple and more user-friendly. Furthermore, our algorithm outperforms the second-best algorithm GNN-RL by $0.14\%$ according to $\Delta$-Acc (ATO $+0.24\%$ vs. GNN-RL $+0.10\%$) when pruning a little more FLOPs (ATO $55.0\%$ vs. GNN-RL $54.0\%$). The gaps between other algorithms and ours are even larger.

\noindent\textbf{MobileNet-V2}. In MobileNet-V2, our method also has good performance. Our algorithm prunes most FLOPs ($45.8\%$) and also achieves the best performance in terms of $\Delta$-ACC ($+0.33$ ).


\subsection{CIFAR-100}
Our comparisons on CIFAR-100 involve ResNet-18 and ResNet-34. All results for the CIFAR-100 dataset are shown in \Cref{tab:1}.

Due to OTOv2 not reporting its result on CIFAR-100 with ResNet-18 and ResNet-34, we produce the results of OTOv2 on CIFAR-100 under the same setting as ours. Compared with results of OTOv2 in \Cref{tab:1}, our algorithm pruned a little more FLOAPs, while our algorithms improved the results largely. 

\subsection{ImageNet}
Subsequently, we employ ATO on ImageNet to demonstrate its effectiveness. In this context, we consider ResNet-34, ResNet-50, and MobileNetV2 as target 
models. The comparison between existing algorithms and ATO is presented in \Cref{tab:2}.

\begin{figure}[!t]
\centering
\includegraphics[width=.4\textwidth]{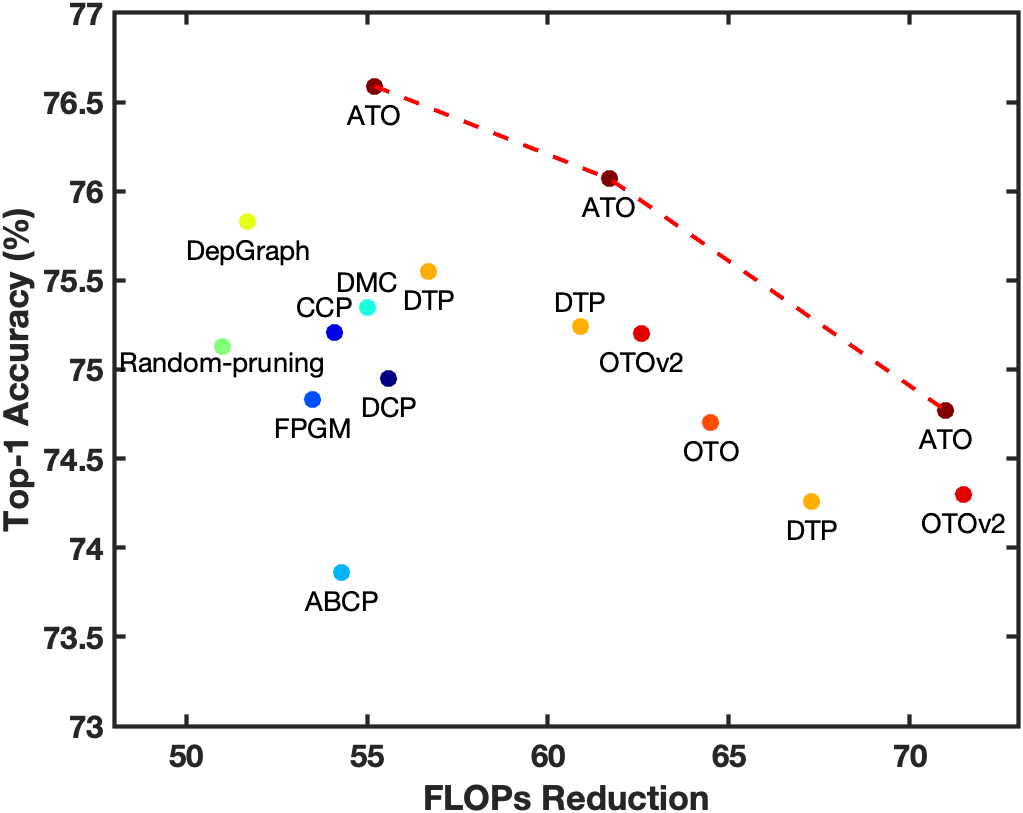}
\caption{ ResNet50 on ImageNet} \label{fig:3}
\end{figure}

\noindent\textbf{ResNet-34}. In the ResNet-34, our algorithm achieves the best performance compared with others under the similar pruned FLOPs although our algorithm has a simpler training procedure.  Our algorithm achieves $72.92\%$ Top-1 accuracy and $91.15\%$ Top-5 accuracy, which are better than other algorithms. SCOP and DMC prune similar FLOPs to our algorithm and have the same baseline results. Nonetheless, our method achieves a pruned Top-1 Accuracy that is $0.30$\% and $0.35$\% superior to SCOP and DMC, respectively. Similar observations are made for Top-5 Accuracy, where our method outperforms SCOP and DMC by $0.17\%$ and $0.04\%$ respectively.

\noindent\textbf{ResNet-50}. In the ResNet-50, we report a performance portfolio under various pruned FLOPs, ranging from $55.2$\% to $71.0$\%. We compare with other counterparts in \Cref{fig:3}. Although increasing pruned FLOPs and parameter reductions typically results in a compromise in accuracy, ATO exhibits a leading edge in terms of top-1 accuracy across various levels of FLOPs reduction. Compared with OTOv2, the results of our algorithm do not suffer from training simplicity. Especially, under pruned FLOPs of $61.7$\%, Top-1 acc of ATO achieves 76.07\%,  which is better than the OTOv2 by $0.87$\% under similar pruned FLOPs. In addition, even when pruned FLOPs is more than $70\%$, ATO still has good performance compared with counterparts.

\noindent\textbf{MobileNetv2}. The lightweight model MobileNetv2 is generally harder to compress. Under the pruned FLOPs of $30\%$, our algorithm archives the best Top-1 acc compared with other methods, even though our algorithm has simpler training procedures. Our algorithm achieves $72.02\%$ Top-1 accuracy and $90.19\%$ Top-5 accuracy while the results of other counterparts are below the baseline results.

\subsection{Ablation Study}
We study the impact of different hyperparameters on model performance. We use the RenNet-56 on CIFAR-10. Note that the falling gap at Epoch 30 is because controller network training starts at Epoch 30 and then we test model performance under the mask vector $\mathbf{w}$, which is equivalent to removing the corresponding ZIGs. 

\noindent\textbf{The impact of $\lambda$.} We study the impact of regularization coefficient $\lambda$ in \cref{eq:1}, and we plot the test accuracy in \Cref{fig:a} and \Cref{fig:e}. From the curves, we can see that $\lambda$ plays an important role in the model training. If the value of $\lambda$ is too small, it will harm model performance, especially when pruned FLOPs are large ($p=0.35$).  

\begin{figure}[H]
  \centering
  \begin{subfigure}{0.44\linewidth}
    \includegraphics[width=.95\textwidth]{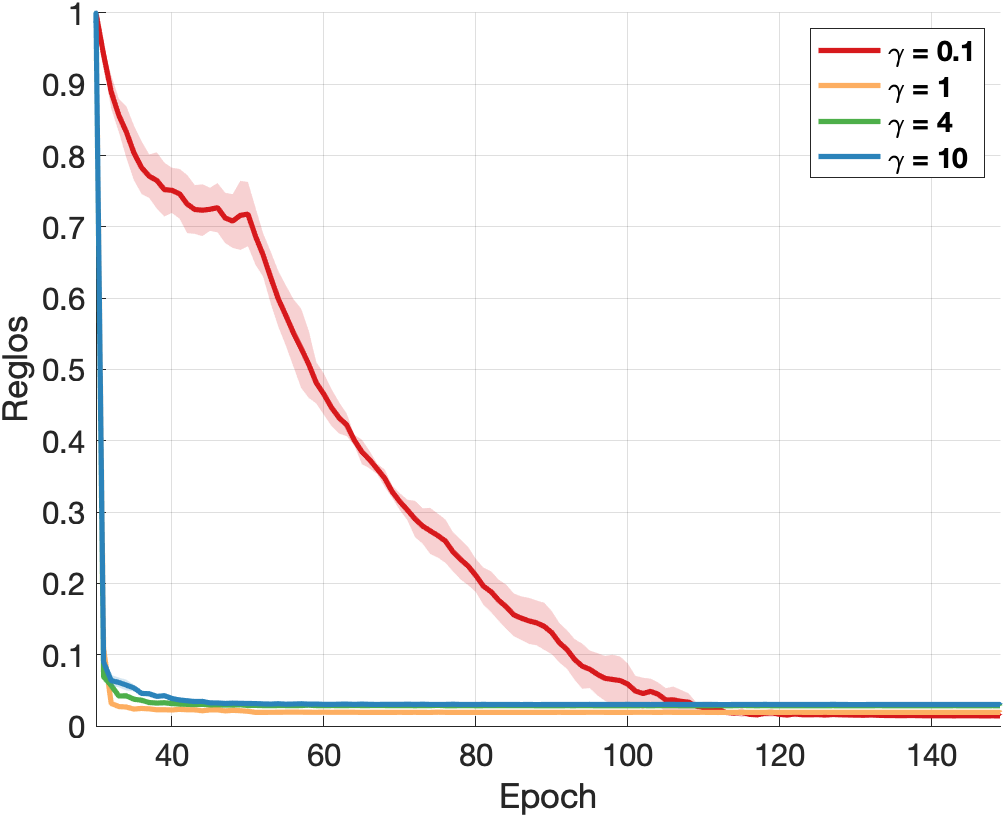}
    \caption{}
  \end{subfigure}
  \begin{subfigure}{0.44\linewidth}
    \includegraphics[width=.95\textwidth]{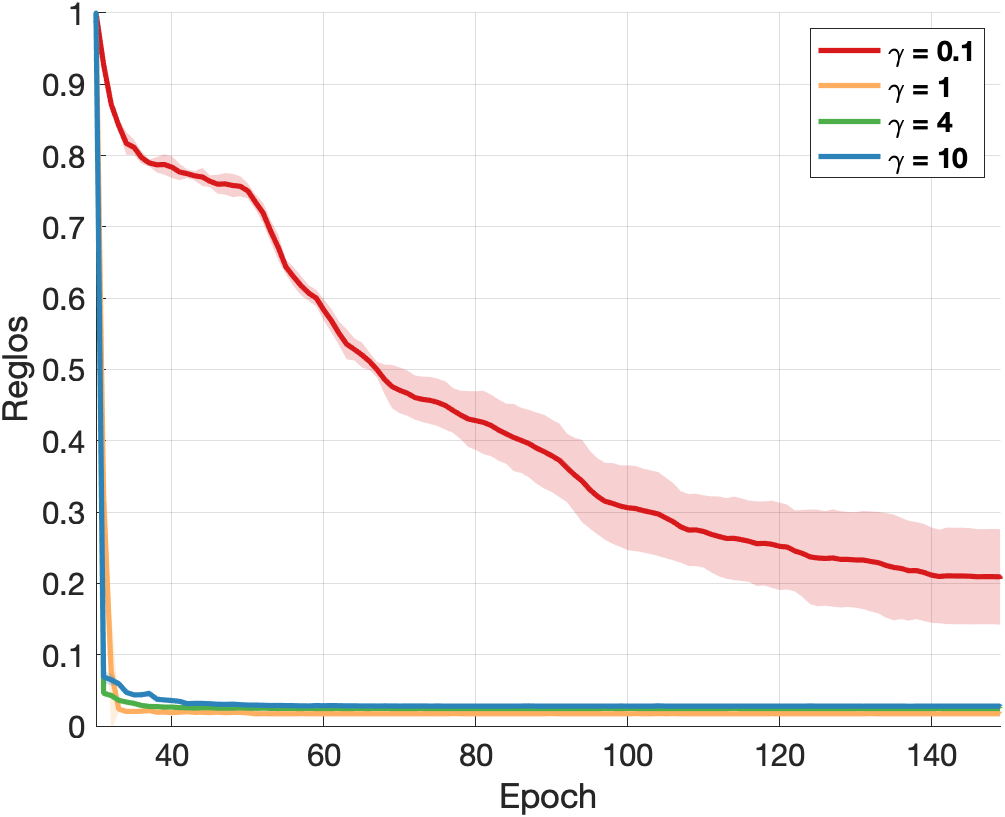}
    \caption{}
  \end{subfigure} 
  \caption{the effect of hyperparameter $\gamma$ in $\mathcal{R}_{\text{FLOPs}}$ in \cref{eq:3}. Experiments are conducted on CIFAR-10 with ResNet-56 with $p=0.45$ (a) and $p=0.35$ (b).}
  \label{fig:regloss}
  \vspace{-5pt}
\end{figure}

\noindent\textbf{The impact of $\gamma$.} We plot the impact of hyper-parameter controlling $\gamma$, which decide the strength of FLOPs regularization  in \cref{eq:3} and we plot the test accuracy in \Cref{fig:b} and \Cref{fig:f}. From the results, we can see that accuracy is lightly affected by $\gamma$ and the curves of different $\gamma$ converge quickly after the model starts training (Epoch = 30). Furthermore, we plot  $\mathcal{R}_{\text{FLOPs}}$ in \Cref{fig:regloss}, and the controller network under different $\gamma$ converges before training stops at Epoch =150 with different speeds. Note the loss value is scaled to [0, 1]. It shows that selecting a too-small $\gamma$ may hinder the controller network from achieving the target FLOPs when the pruning rate is large. Otherwise, the training of the controller network is robust.

\noindent\textbf{The impact of $T_w$.} We study the impact of $T_w$ when training target model in \Cref{fig:c} and \Cref{fig:g}. Since we start training the controller network at Epoch 30 and we need a mask $\mathbf{w}$ from the controller network, $T_w$ is selected from set \{30, 50, 100, 150\}. In general, they can converge to the optimal under the mask with different  $T_w$. 

\noindent\textbf{The impact of projection operator.} To verify our proposed is algorithm flexible to the projection operator, we plot the test accuracy of the proximal gradient projector in \ref{eq:2_2} and Half-Space projector \ref{eq:2_1}. In OTO \cite{oto} and OTOv2 \cite{chen2023otov2}, due to the limitation of the manually selected static mask, they have to use (D)HSPG \cite{chen2023otov2} to achieve better performance. In the test, we show that the controller network helps us solve these tough questions and there is no big difference between the two projection operators. Due to the easy implementation and the better efficiency of the proximal gradient projector, we use it in our experiments.

\section{Conclusion}
In this paper, we investigate automatic network pruning from scratch and address the limitations found in existing algorithms, including 1) complex multi-step training procedures and 2) the suboptimal outcomes associated with statically chosen pruning groups. Our solution, Auto-Train-Once (ATO), introduces an innovative network pruning algorithm designed to automatically reduce the computational and storage costs of DNNs without reliance on the extra fine-tuning step. During the model training phase, a controller network dynamically generates the binary mask to guide the pruning of the target model. Additionally, we have developed a novel stochastic gradient algorithm that offers flexibility in the choice of projection operators and enhances the coordination between the model training and the controller network training, thereby improving pruning performance. In addition, we present a theoretical analysis under mild assumptions to guarantee convergence, along with extensive experiments. The experiment results demonstrate that our algorithm achieves state-of-the-art performance across various model architectures, including ResNet18, ResNet34, ResNet50, ResNet56, and MobileNetv2 on standard benchmark datasets such as CIFAR-10, CIFAR-100, as well as ImageNet.

{
    \small
    \bibliographystyle{ieeenat_fullname}
    \bibliography{main}
}


\newpage
\onecolumn
\section{Implementation Details}
In this section, we provide more details of the implementation. 

We follow the standard training example in the training on CIFAR-10 and CIFAR-100 datasets. The model is trained for 300 epochs. We select SGD as the optimizer with a learning rate of $0.1$, a momentum of $0.9$, and a weight decay of $10^{-4}$. The value of $p$ in \cref{eq:3} for all datasets and models are presented in \Cref{tab:5}.

\begin{table} [th]
  \centering
  \caption{Choice of $p$ in \cref{eq:3} for different datasets}
  \label{tab:5}
    \vspace*{-6pt}
\begin{tabular}{ccc}
\hline DataSet & Model & $p$ \\
\hline 
\multirow{3}{*}{CIFAR-10} & ResNet-18 & 0.1908 \\
 & ResNet-56 &  0.3500 / 0.4500\\
 & MobileNetV2 & 0.4900\\
\hline
\multirow{2}{*}{CIFAR-100} &ResNet-18 & 0.5952\\
& ResNet-34 & 0.5020\\ 
\hline 
\multirow{3}{*}{ImageNet} & ResNet-34 & 0.5400\\
& ResNet-50 & 0.3700/0.2960/0.1930\\
& MobileNetV2 & 0.6578\\
\hline 
\end{tabular}
\end{table}

To train ResNet models on ImageNet, we follow the ImageNet training setting \footnote{https://github.com/tianyic/only\_train\_once/blob/main/tutorials/02.resnet50\_imagenet.ipynb} in OTOv2 \cite{chen2023otov2} and train the ResNet models for 240 epochs. For MobileNet-V2, we train the model for 300 epochs with the cos-annealing learning rate scheduler 
and a start learning rate of 0.05, and weight decay $4\times10^{-5}$ as mentioned in their original paper \cite{sandler2018mobilenetv2}.
As mentioned in the paper, we set $T_{start}$ to around $10\%$ of the total training epochs and $T_{end}$ to $50\%$ of the total training epochs.

\begin{table} [th]
  \centering
  \caption{ Architecture of Controller Network}
  \label{tab:5}
    \vspace*{-6pt}
\begin{tabular}{lc}
\hline Layer Type & Shape \\
\hline 
Input & $|\mathcal{B}| \times 64$ \\
Bi-GRU  & $64 \times 128 \times 2$ \\
LayerNorm + ReLU & 256 \\
Linear Layer & $256 \times |B_i|, B_i \subset \mathcal{B} $  \\
Concatenate & $|\mathcal{G}|$\\
Gumbel-Sigmoid  & - \\
Round $\rightarrow \mathbf{w}$& - \\
\hline
\end{tabular}
\end{table}

\Cref{tab:5} shows the architecture of the controller network. Controller Network uses bi-directional gated recurrent units (GRU) \cite{cho2014properties} followed by linear layers with output as ${o}$, where the dimension of ${o}$ is equal to the size of ZIGs $\mathcal{G}$. To reduce the training costs, we divide ZIGs $\mathcal{G}$ into multiple disjoint blocks $\mathcal{B} =\{B_1, \cdots, B_{|\mathcal{B}|}\}$ ($B_1 \cup B_2 \cup \cdots B_{|\mathcal{B}|}  = \mathcal{G}$ and $\sum |B_i| = |\mathcal{G}|$), and each block is one layer in ResNet models or one InvertedResidual block in MobileNetv2. Finally, the linear layer projects the output to the $|B_i|$ and $\sum |B_i| = |\mathcal{G}|$. 

The model mask vector is generated as below:
\begin{align}
\mathbf{w} = \text{round}(\text{sigmoid}(({o} + s + b)/\tau)) \nonumber
\end{align}
where $\text{sigmoid}(\cdot)$ is the sigmoid function, $\text{round}(\cdot)$ is the rounding function, $s$ is sampled from Gumbel distribution ($s\sim\text{Gumbel}(0,1)$), $b$ and $\tau$ are constants with values as $3.0$ and $0.4$.

\section{Theoretical analysis}
In this section, we provide a detailed convergence analysis of our algorithms. We define the $z$ as the vector of the target model weight and we reformulated our algorithm (i.e., ATO) in the mirror descent format, as presented in \cref{alg:3}, to facilitate analysis. It should be noted that $t$ in \cref{alg:3} denotes the update iteration steps to discuss the training progression in each step, instead of the training epoch index as before. $g(z) = \sum_{g \in \mathcal{G}} \lambda_g\left\|[\mathcal{M}]_g\right\|$ and $g(z)$ varies as training. In the theoretical analysis, we assume the change of $g(z)$ is slow and regard it as static. 

As discussed in the paper, we define a Bregman distance \cite{ghadimi2016mini, huang2021efficient, bao2022doubly} associated with $\phi(z)$ as follows:
\begin{align}
 D_{\phi}(z^{\prime},z) = \phi(z^{\prime}) - \big[\phi(z) + \langle\nabla \phi(z), z^{\prime}-z\rangle\big], \quad \forall{z,z^{\prime}}
\end{align}
Given that $\phi(z)=\frac{1}{2}z^T A z $ in our paper, we have $D_{\phi}(z^{\prime},z) = \frac{1}{2}(z^{\prime} - z)^T A_t(z^{\prime} - z)$.
For the non-adaptive optimizer (e.g., SGD), we choose $A = I$. For adaptive optimizer (i.e., Adam), we update $A_t$ as in Adam-type algorithms \cite{kingma2014adam}, defined as
\begin{align} 
 & \tilde{v}_0=0, \ \tilde{v}_t = \beta \tilde{v}_{t-1} + (1 - \beta) \nabla_z \mathcal{L}(z_t;\xi_t)^2, \nonumber\\
 &A_t= \mbox{diag}( \sqrt{\tilde{v}_t} + \epsilon)\nonumber
\end{align}
where $\tilde{v}$ is the second-moment estimator, and $\epsilon$ is a term to improve numerical stability in Adam-type optimizer. 
Therefore, $\phi(z)$ is a $\epsilon$-strongly convex function. We first give some useful assumptions and lemmas.

\begin{assumption} \label{ass:1} (Unbiased gradient and bounded variance) Set $\xi = (x,y)$ as a batch of samples, $z$ as the vector of network weight $\mathcal{M}$ and $\mathcal{J}(z) = \mathcal{L}(z) + g(z)$. Assume loss function $\mathcal{L}(z;\xi)$ with samples $\xi$ has an unbiased stochastic gradient with bounded variance $\sigma^2$, i.e., 
\begin{align}
&\mathbb{E}[\nabla_z \mathcal{L}(z;\xi)] = \nabla_{z} \mathcal{L}(z) \\
&\mathbb{E}\| \nabla_{z} \mathcal{L}(z; \xi) - \nabla_{z} \mathcal{L}(z)\|^2 \leq \sigma^2
\end{align}
\end{assumption}

\begin{assumption} \label{ass:2} (Gradient Lipschitz)
The loss function $\mathcal{L}(\mathcal{M})$ has a $L$-Lipschitz gradient, i.e.,
 for $\forall z_1, z_2$ two different network weights, we have
\begin{align}
& \|\nabla_{z} \mathcal{L}(z_1) - \nabla_{z} \mathcal{L}(z_2) \| \leq L \|z_1 - z_2\|
\end{align}
\end{assumption}
Assumption \ref{ass:1} and  \ref{ass:2} are standard assumptions in stochastic optimization and are widely used in deep learning convergence analysis \cite{ghadimi2016mini, bao2020fast, bao2022accelerated, mei2024bayesian, mei2024projection, wu2024solving, wu2023faster}.

\begin{lemma} \label{lem:A1}
(Lemma 1 in \cite{ghadimi2016mini}) Assume $g(z)$  is a convex and possibly nonsmooth function.
Let $z^+_{t+1} = \arg\min_{z} \big\{\langle z, \nabla_z \mathcal{L}(z) \rangle  + \frac{1}{\eta}D_{\phi_t}(z, z_t)+ g(z)\big\} $ and $\mathcal{P}_t  = \frac{1}{\eta}(z_t - z^+_{t+1})$. Then we have
\begin{align}
 \langle \nabla_z \mathcal{L}(z),  \mathcal{P}_t \rangle \geq \epsilon \|\mathcal{P}_t \|^2 + \frac{1}{\eta} \big[g(z^+_{t+1}) - g(z_t)\big]
\end{align}
where $\epsilon>0$ depends on $\epsilon$-strongly convex function $\phi_t(z)$.
\end{lemma}
Based on Lemma \ref{lem:A1}, 
let $z_{t+1} = \arg\min_{z} \big\{\langle z, m_t \rangle  + \frac{1}{\eta}D_{\phi_t}(z, z_t)+ g(z)\big\}$, and define $\Tilde{\mathcal{P}}_t = \frac{1}{\eta} [z_t - z_{t+1}]$. We have
\begin{align} \label{eq:14}
 \langle m_t, \Tilde{\mathcal{P}}_t\rangle \geq \epsilon \|\Tilde{\mathcal{P}}_t\|^2 + \frac{1}{\eta} (g(z_{t+1}) - g(x_{t})) 
\end{align}

\begin{algorithm}[tb]
\caption{ATO Algorithm (Mirror Descant) }
\label{alg:3}
\begin{algorithmic}[1] 
\STATE {\bfseries Input:} Target model with model weights vector $z$ (no need to be pre-tained). Datasets $D$, $D_{\text{CN}}$, $\gamma$, $\lambda$, $\eta$, total training iteration $T$ and each epoch of $\mathcal{D}$ has $Q$ iterations; controller network training steps $T_{\text{start}}$ and $T_{\text{end}}$.\\
\FOR{$t = 1, 2 \ldots, T$}
\STATE Sample $\xi = (x,y)$ and compute the stochastic gradient $\nabla \mathcal{L}(z;\xi)$ \\
\STATE Compute gradient estimator
$m_t = (1 - \alpha_t) m_{t-1} + \alpha_t \nabla \mathcal{L}(z;\xi)$ \\
\STATE  Update model weight $z_{t+1} = \arg\min_{z} \big\{\langle m_t, z\rangle + \frac{1}{\eta_t}D_{\phi_t}(z, z_t) + g(z) \big\} $; \\
\IF {$T_{\text{start}} Q \leq T  \leq T_{\text{end}} Q$} 
\STATE  $\mathcal{W}, \mathbf{w} \leftarrow \text{CN-Update}(\mathcal{M}, \mathcal{W}, \mathbf{w}, D_{\text{C}})$\\
\ENDIF
\ENDFOR
\STATE {\bfseries Output:} Directly remove pruned structures and construct a slimmer model.
\end{algorithmic}
\end{algorithm}

\begin{lemma} \label{lem:A2}
 Assume that the stochastic partial derivatives $m_{t+1}$ be generated from Algorithm \ref{alg:1}, we have
 \begin{align}
 \mathbb{E}\|\nabla_{z} \mathcal{L}(z_{t+1}) - m_{t+1}\|^2
 & \leq (1-\alpha_{t+1}) \mathbb{E} \|\nabla_{z} \mathcal{L}(z_t) ) - m_{t}\|^2 + \frac{ L^2\eta_t^2}{\alpha_{t+1}} \mathbb{E}\|\tilde{\mathcal{P}_t}\|^2 + \frac{\alpha_{t+1}^2\sigma^2}{b}
 \end{align}
\end{lemma}

\begin{proof}
Since $m_{t+1} =  \alpha_{t+1}\nabla_{z} \mathcal{L}(z_{t+1};\xi_{t+1}) + (1-\alpha_{t+1})m_{t}$, we have
\begin{align}
 &\mathbb{E}\|\nabla_{z} \mathcal{L}(z_{t+1}) - m_{t+1}\|^2  =  \mathbb{E}\|\nabla_{z} \mathcal{L}(z_{t+1}) -  \alpha_{t+1}\nabla_{z} \mathcal{L}(z_{t+1};\xi_{t+1}) - (1 - \alpha_{t+1})m_{t} \|^2 \nonumber \\
 =& \mathbb{E}\|\alpha_{t+1}(\nabla_{z} \mathcal{L}(z_{t+1}) - \nabla_{z} \mathcal{L}(z_{t+1};\xi_{t+1})) + (1-\alpha_{t+1})(\nabla_{z} \mathcal{L}(z_t) - m_{t})\nonumber \\
 & \quad + (1-\alpha_{t+1})\big( \nabla_{z} \mathcal{L}(z_{t+1} ) - \nabla_{z} \mathcal{L}(z_t) \big)\|^2 \nonumber \\
 \stackrel{(a)}{=} & \mathbb{E}\| (1-\alpha_{t+1})(\nabla_{z} \mathcal{L}(z_t) - m_{t}) + (1-\alpha_{t+1})\big( \nabla_{z} \mathcal{L}(z_{t+1} ) - \nabla_{z} \mathcal{L}(z_t) )\|^2 \nonumber \\
 & \quad + \alpha^2_{t+1}\mathbb{E}\|\nabla_{z} \mathcal{L}(z_{t+1} ) - \nabla_{z} \mathcal{L}(z_{t+1} ;\xi_{t+1})\|^2 \nonumber \\
 \leq & (1-\alpha_{t+1})^2(1+\frac{1}{\alpha_{t+1}})\mathbb{E} \|\nabla_{z} \mathcal{L}(z_{t+1} ) - \nabla_{z} \mathcal{L}(z_t)\|^2 \nonumber \\
 & \quad +(1-\alpha_{t+1})^2(1+\alpha_{t+1})\mathbb{E} \|\nabla_{z} \mathcal{L}(z_t) - m_{t}\|^2+ \alpha^2_{t+1}\mathbb{E}\|\nabla_{z} \mathcal{L}(z_{t+1} ) - \nabla_{z} \mathcal{L}(z_{t+1} ;\xi_{t+1})\|^2 \nonumber \\
 \stackrel{(b)}{\leq} & (1-\alpha_{t+1})\mathbb{E} \|\nabla_{z} \mathcal{L}(z_t) - m_{t}\|^2 + \frac{1}{\alpha_{t+1}}\mathbb{E}\|\nabla_{z} \mathcal{L}(z_{t+1} ) - \nabla_{z} \mathcal{L}(z_t) )\|^2  + \frac{\alpha^2_{t+1}\sigma^2}{b} \nonumber \\
 \leq & (1-\alpha_{t+1}) \mathbb{E} \|\nabla_{z} \mathcal{L}(z_t) - m_{t}\|^2 + \frac{ L^2 }{\alpha_{t+1}} \mathbb{E}\|z_{t+1} - z_t\|^2 + \frac{\alpha_{t+1}^2\sigma^2}{b} \nonumber \\
 = & (1-\alpha_{t+1}) \mathbb{E} \|\nabla_{z} \mathcal{L}(z_t) - m_{t}\|^2 + \frac{ L^2\eta_t^2}{\alpha_{t+1}} \mathbb{E}\|\tilde{\mathcal{P}}_t\|^2 + \frac{\alpha_{t+1}^2\sigma^2}{b} \nonumber
\end{align}
where the (a) is due to $\mathbb{E}_{\xi_{t+1}}[\nabla \mathcal{L}(z_{t+1} ;\xi_{t+1})]=\nabla \mathcal{L}(z_{t+1} )$; the (b) holds by $0\leq \alpha_{t+1} \leq 1$ such that  $(1-\alpha_{t+1})^2(1+\alpha_{t+1})=1-\alpha_{t+1}-\alpha_{t+1}^2+
\alpha_{t+1}^3\leq 1-\alpha_{t+1}$ and $(1-\alpha_{t+1})^2(1+\frac{1}{\alpha_{t+1}}) \leq (1-\alpha_{t+1})(1+\frac{1}{\alpha_{t+1}})  =-\alpha_{t+1}+\frac{1}{\alpha_{t+1}}\leq \frac{1}{\alpha_{t+1}}$, and the $b$ is the batch size; the last inequality holds by the definition of $\tilde{\mathcal{P}}_t$ in \cref{eq:14}
\end{proof}

\begin{lemma} 
Let $z_{t+1}=\arg \min _{z  }\left\{\left\langle m_t, z\right\rangle+\frac{1}{\eta_t} D_{\phi_t}\left(z, z_t\right)+g(z)\right\}$ and $z_{t+1}^{+}=\arg \min _{z  }\left\{\left\langle\nabla \mathcal{L}\left(z_t\right), z\right\rangle+\right.$ $\left.\frac{1}{\eta_t} D_{\phi_t}\left(z, z_t\right) + g(z)\right\}$, we have
\begin{align}
\left\|\nabla \mathcal{L}\left(z_t\right)-m_t\right\| \geq \epsilon\left\|\mathcal{P}_t-\tilde{\mathcal{P}}_t\right\|
\end{align}  \label{lem:A3}
\end{lemma} 
\begin{proof}
based on the definition of $z_{t+1}$ and $z_t$, and the convex property, we have, 
\begin{align}
& \left\langle m_t + \nabla g\left(z_{t+1}\right) + \frac{1}{\eta_t}\left(\nabla \mathcal{L}_t\left(z_{t+1}\right)-\nabla \mathcal{L}_t\left(z_t\right)\right), z - z_{t+1}\right\rangle \geq 0  \label{eq:17}\\
& \left\langle\nabla \mathcal{L}\left(z_t\right)+\nabla g\left(z_{t+1}^{+}\right)+\frac{1}{\eta_t}\left(\nabla \mathcal{L}_t\left(z_{t+1}^{+}\right)-\nabla \mathcal{L}_t\left(z_t\right)\right), z-z_{t+1}^{+}\right\rangle \geq 0 \label{eq:18}
\end{align}

where $\nabla g\left(z_{t+1}\right) \in \partial g\left(z_{t+1}\right)$. Taking $z=z_{t+1}^{+}$in the \cref{eq:17} and $z=z_{t+1}$ in the \cref{eq:18}, by the convexity of $g(x)$, we have

\begin{align}
\left\langle m_t, z_{t+1}^{+}-z_{t+1}\right\rangle & \geq\left\langle\nabla g\left(z_{t+1}\right), z_{t+1}-z_{t+1}^{+}\right\rangle+\frac{1}{\eta_t}\left\langle\nabla \mathcal{L}_t\left(z_{t+1}\right)-\nabla \mathcal{L}_t\left(z_t\right), z_{t+1} - z_{t+1}^{+}\right\rangle \nonumber\\
& \geq g\left(z_{t+1}\right)-g\left(z_{t+1}^{+}\right)+\frac{1}{\eta_t}\left\langle\nabla \mathcal{L}_t\left(z_{t+1}\right)-\nabla \mathcal{L}_t\left(z_t\right), z_{t+1}-z_{t+1}^{+}\right\rangle \\
\left\langle\nabla \mathcal{L}\left(z_t\right), z_{t+1}-z_{t+1}^{+}\right\rangle & \geq\left\langle\nabla g\left(z_{t+1}^{+}\right), z_{t+1}^{+}-z_{t+1}\right\rangle+\frac{1}{\eta_t}\left\langle\nabla \mathcal{L}_t\left(z_{t+1}^{+}\right)-\nabla \mathcal{L}_t\left(z_t\right), z_{t+1}^{+}-z_{t+1}\right\rangle \nonumber \\
& \geq g\left(z_{t+1}^{+}\right)-g\left(z_{t+1}\right)+\frac{1}{\eta_t}\left\langle\nabla \mathcal{L}_t\left(z_{t+1}^{+}\right)-\nabla \mathcal{L}_t\left(z_t\right), z_{t+1}^{+}-z_{t+1}\right\rangle
\end{align}

Summing up the above inequalities, we obtain
\begin{align}
\left\langle\nabla \mathcal{L} \left(z_t\right)-m_t, z_{t+1}-z_{t+1}^{+}\right\rangle & \geq \frac{1}{\eta_t}\left\langle\nabla \mathcal{L}_t\left(z_{t+1}^{+}\right)-\nabla \mathcal{L}_t\left(z_{t+1}\right), z_{t+1}^{+}-z_{t+1}\right\rangle \geq \frac{\epsilon}{\eta_t}\left\|z_{t+1}^{+}-z_{t+1}\right\|^2 \nonumber
\end{align}
where the last inequality is due to the $\epsilon$-strongly convex function $\mathcal{L}_t(z)$.
Since $\left\|\nabla \mathcal{L}\left(z_t\right)-m_t\right\|\left\|z_{t+1}-z_{t+1}^{+}\right\| \geq \left\langle\nabla \mathcal{L}\left(z_t\right) - m_t, z_{t+1}-z_{t+1}^{+}\right\rangle$ and $\left\|\mathcal{P}_t-\tilde{\mathcal{P}}_t\right\|=\| \frac{1}{\eta_t}\left(z_t-z_{t+1}^{+}\right)-$ $\frac{1}{\eta_t}\left(z_t-z_{t+1}\right)\left\|=\frac{1}{\eta_t}\right\| z_{t+1}^{+}-z_{t+1} \|$, we have
\begin{align}
\left\|\nabla \mathcal{L}\left(z_t\right)-m_t\right\| \geq \epsilon\left\|\mathcal{P}_t-\tilde{\mathcal{P}}_t\right\|
\end{align}
\end{proof}
\begin{lemma} \label{lem:A4}
Suppose the sequence $\{z_t \}_{t=1}^T$ be generated from Algorithms \ref{alg:1}. Let $0<\eta_t \leq \min \{1, \frac{\epsilon}{4L } \}$,
we have
\begin{align}
\mathcal{J}\left(z_{t+1}\right) & \leq \mathcal{J}\left(z_t\right)-\frac{ \eta_t \epsilon}{4}\left\|\mathcal{P}_t\right\|^2 +\frac{2\eta_t}{\epsilon}\left\|m_t - \nabla_z \mathcal{L} \left(z_t\right)\right\|^2 
\end{align}
\end{lemma}

\begin{proof}

Since $z_{t+1}=\arg \min _{z  }\left\{\left\langle m_t, z\right\rangle+\frac{1}{\eta_t} D_{\phi_t}\left(z, z_t\right)+g(z)\right\}$ and $\Tilde{\mathcal{P}}_t = \frac{1}{\eta} (z_t - z_{t+1})$, and function $\mathcal{L}(z)$ has $L$-Lipschitz continuous gradient. we have
\begin{align}
\mathcal{L}\left(z_{t+1}\right) & \leq \mathcal{L}\left(z_t\right)+\left\langle\nabla \mathcal{L}\left(z_t\right), z_{t+1} - z_t\right\rangle+\frac{L}{2}\left\|z_{t+1}-z_t\right\|^2 \nonumber\\
& =\mathcal{L}\left(z_t\right)-\eta_t\left\langle\nabla \mathcal{L}\left(z_t\right), \tilde{\mathcal{P}_t}\right\rangle+\frac{\eta_t^2 L}{2}\left\|\tilde{\mathcal{P}}_t\right\|^2 \nonumber\\
& =\mathcal{L}\left(z_t\right)-\eta_t\left\langle m_t, \tilde{\mathcal{P}} t\right\rangle+\eta_t\left\langle m_t-\nabla \mathcal{L}\left(z_t\right), \tilde{\mathcal{P}_t}\right\rangle+\frac{\eta_t^2 L}{2}\left\|\tilde{\mathcal{P}}_t\right\|^2 \nonumber\\
& \stackrel{(a)}{\leq} \mathcal{L}\left(z_t\right)-\eta_t \epsilon\left\|\tilde{\mathcal{P}}_t\right\|^2-g\left(z_{t+1}\right)+g\left(z_t\right)+\eta_t\left\langle m_t-\nabla \mathcal{L}\left(z_t\right), \tilde{\mathcal{P}}_t\right\rangle+\frac{\eta_t^2 L}{2}\left\|\tilde{\mathcal{P}}_t\right\|^2 \nonumber \\
& \stackrel{(b)}{\leq }   \mathcal{L}\left(z_t\right)+\left(\frac{\eta_t^2 L}{2}-\frac{3 \eta_t \epsilon}{4}\right)\left\|\tilde{\mathcal{P}}_t\right\|^2-g\left(z_{t+1}\right)+g\left(z_t\right)+\frac{\eta_t}{\epsilon}\left\|m_t-\nabla \mathcal{L}\left(z_t\right)\right\|^2
\end{align}
where the (a)holds by the above Lemma \ref{lem:A1}, and the (b) holds by the following Cauchy-Schwarz inequality and Young’s inequality as
\begin{align}
\left\langle m_t-\nabla \mathcal{L}\left(z_t\right), \tilde{\mathcal{P}_t}\right\rangle & \leq\left\|m_t-\nabla \mathcal{L}\left(z_t\right)\right\|\left\|\tilde{\mathcal{P}_t}\right\| \nonumber\\
& \leq \frac{1}{\epsilon}\left\|m_t-\nabla \mathcal{L}\left(z_t\right)\right\|^2+\frac{\epsilon}{4}\left\|\tilde{\mathcal{P}_t}\right\|^2 \nonumber
\end{align}
Since $\mathcal{J}(z) = \mathcal{L}(z) + g(z)$, we have
\begin{align}
\mathcal{J}\left(z_{t+1}\right) & \leq \mathcal{J}\left(z_t\right)+\left(\frac{\eta_t^2 L}{2}-\frac{3 \eta_t \epsilon}{4}\right)\left\|\tilde{\mathcal{P}}_t\right\|^2+\frac{\eta_t}{\epsilon}\left\|m_t - \nabla_z \mathcal{L} \left(z_t\right)\right\|^2 \nonumber \\
& \leq \mathcal{J}\left(z_t\right)-\frac{5 \eta_t \epsilon}{8}\left\|\tilde{\mathcal{P}}_t\right\|^2 +\frac{\eta_t}{\epsilon}\left\|m_t - \nabla_z \mathcal{L} \left(z_t\right)\right\|^2 
\end{align}

where the last inequality is due to $0< \eta_t \leq \frac{\epsilon}{4L}$. Based on \cref{lem:A3}, we have
\begin{align}
\left\|\mathcal{P}_t\right\|^2  \leq 2\left\|\tilde{\mathcal{P}}_t\right\|^2+2\left\|\tilde{\mathcal{P}}_t-\mathcal{P}_t\right\|^2  \leq 2\left\|\tilde{\mathcal{P}}_t\right\|^2+\frac{2}{\epsilon^2}\left\|m_t - \nabla \mathcal{L} \left(z_t\right)\right\|^2 
\end{align}
Finally, we have
\begin{align}
\mathcal{J}\left(z_{t+1}\right)  \leq \mathcal{J}\left(z_t\right)-\frac{ \eta_t \epsilon}{8}\left\|\tilde{\mathcal{P}}_t\right\|^2 -\frac{ \eta_t \epsilon}{4}\left\|\mathcal{P}_t\right\|^2  + \frac{2 \eta_t}{\epsilon}\left\|m_t - \nabla_z \mathcal{L} \left(z_t\right)\right\|^2 
\end{align}

\end{proof}

\begin{theorem} 
(Restatement of Theorem 1)
Assume that the sequence $\{z_t \}_{t=1}^T$ be generated from the Algorithm \ref{alg:1}. When  we have 
$\eta_t = \frac{\hat{c}}{(\bar{c}+t)^{1/2}}$, $ \frac{\hat{c}}{\bar{c}^{1/2}} \leq \min \{1, \frac{\epsilon}{4L }\}$, $c_1 = \frac{4L}{\epsilon }, \alpha_{t+1} = c_1\eta_t$, we have
\begin{align} 
 \frac{1}{T} \sum_{t=1}^T \mathbb{E} \|\mathcal{P}_{t}\|   \leq \frac{\sqrt{G}\bar{c}^{1/4}}{T^{1/2}} + \frac{\sqrt{G}}{T^{1/4}}
\end{align}
where $G =  \frac{4 (\mathcal{J}(z_1) - \mathcal{J}(z^*))}{\epsilon \hat{c}} + \frac{2\sigma^2}{b L \epsilon \hat{c}} + \frac{2\bar{c}\sigma^2}{\hat{c} \epsilon L b}\ln(\bar{c}+T) $.
\end{theorem}

\begin{proof}
$\eta_t = \frac{\hat{c}}{(\bar{c}+t)^{1/2}}$ on $t$ is decreasing, $\eta_t \leq \eta_0 = \frac{\hat{c}}{\bar{c}^{1/2}} \leq \min \{1, \frac{\epsilon}{4L }\}$ for any $t\geq 0$. At the same time, $c_1 = \frac{4L}{\epsilon }$.
We have $\alpha_{t+1} = c_1\eta_t \leq \frac{c_1\hat{c}}{\bar{c}^{1/2}}\leq 1$.

According to Lemma \ref{lem:A2}, we have
 \begin{align} \label{eq:H1}
  & \mathbb{E} \|\nabla_{z} \mathcal{L}(z_{t+1} ) - m_{t+1}\|^2 - \mathbb{E} \|\nabla_{z} \mathcal{L}(z_t) - m_{t}\|^2  \\
 \leq & -\alpha_{t+1}\mathbb{E} \|\nabla_{z} \mathcal{L}(z_t) -m_{t}\|^2 + L^2\eta^2_t/\alpha_{t+1}\mathbb{E} \|\tilde{\mathcal{P}}_{t} \|^2 + \frac{\alpha_{t+1}^2\sigma^2}{b}  \nonumber \\
 =& - c_1 \eta_t\mathbb{E} \|\nabla_{z} \mathcal{L}(z_t) - m_{t}\|^2 + L^2 \eta_t/c_1\mathbb{E} \|\tilde{\mathcal{P}}_{t}\|^2 + \frac{c_1^2\eta_t^2\sigma^2}{b} \nonumber \\
\leq &  -  \frac{4 L \eta_t}{\epsilon}  \mathbb{E} \|\nabla_{z} \mathcal{L}(z_t) - m_t\|^2 + \frac{\epsilon L \eta_t}{4}  \mathbb{E} \|\tilde{\mathcal{P}}_{t}\|^2 + \frac{\bar{c}\eta^2_t\sigma^2}{\hat{c}^2q} \nonumber
 \end{align}
 where the above equality holds by $\alpha_{t+1}=c_1\eta_t$, and the last inequality is due to $c_1 = \frac{4L}{ \epsilon} $.

According to Lemma \ref{lem:A4}, we have
\begin{align}
\mathcal{J}\left(z_{t+1}\right)  \leq \mathcal{J}\left(z_t\right) -\frac{ \eta_t \epsilon}{8}\left\|\tilde{\mathcal{P}}_t\right\|^2 -\frac{ \eta_t \epsilon}{4}\left\|\mathcal{P}_t\right\|^2  + \frac{2 \eta_t}{\epsilon}\left\|m_t - \nabla_z \mathcal{L} \left(z_t\right)\right\|^2 
\end{align}

Next, we define a \emph{Lyapunov} function, for any $t\geq 1$
 \begin{align}
 \Omega_t & = \mathbb{E}\big [\mathcal{J}(z_t) + \frac{1}{2L}  \|\nabla_{z} \mathcal{L}(z_t)  - m_{t}\|^2] \nonumber
 \end{align}
Then we have
 \begin{align}
 & \Omega_{t+1} - \Omega_t \nonumber = \mathbb{E}\big[\mathcal{J}(z_{t+1}) - \mathcal{J}(z_t)\big]
 + \frac{1}{2} [ \mathbb{E}\|\nabla_{z} \mathcal{L}(z_{t+1} )-m_{t+1}\|^2 \nonumber - \mathbb{E}\|\nabla_{z} \mathcal{L}(z_t) - m_{t}\|^2 ]\nonumber \\
\leq & -\frac{ \eta_t \epsilon}{8}\left\|\tilde{\mathcal{P}}_t\right\|^2 -\frac{ \eta_t \epsilon}{4}\left\|\mathcal{P}_t\right\|^2  + \frac{2 \eta_t}{\epsilon}\left\|m_t - \nabla_z \mathcal{L} \left(z_t\right)\right\|^2  + \frac{1}{2L } \bigg( -  \frac{4 L \eta_t}{\epsilon}  \mathbb{E} \|\nabla_{z} \mathcal{L}(z_t) - m_t\|^2 + \frac{\epsilon L \eta_t}{4}  \mathbb{E} \|\tilde{\mathcal{P}}_{t}\|^2 + \frac{\bar{c}\eta^2_t\sigma^2}{\hat{c}^2q} \bigg) \nonumber\\
\leq &  - \frac{\epsilon\eta_t}{4}\mathbb{E}\|\mathcal{P}_{t}\|^2 + \frac{\bar{c} \sigma^2}{2 \hat{c}^2 L q}\eta^2_t \nonumber
 \end{align}

Then we have
\begin{align} \label{eq:31}
\eta_t \mathbb{E}\|\mathcal{P}_{t}\|^2 \leq \frac{4(\Omega_t - \Omega_{t+1})}{\epsilon} + \frac{2 \bar{c} \sigma^2}{\epsilon \hat{c}^2 L b}\eta^2_t
\end{align}
Taking average over $t=1,2,\cdots,T$ on both sides of \eqref{eq:31}, we have
\begin{align}
 & \frac{1}{T} \sum_{t=1}^T   \eta_t \mathbb{E}\|\mathcal{P}_{t}\|^2   \leq  \sum_{t=1}^T \frac{4(\Omega_t - \Omega_{t+1})}{T\epsilon} + \frac{1}{T}\sum_{t=1}^T\frac{2\bar{c}\sigma^2}{ \epsilon L \hat{c}^2 b}\eta^2_t.
\end{align}
In addition, we have
\begin{align} \label{eq:H6}
 \Omega_1 = \mathcal{J}(z_1) + \frac{1}{2L }  \mathbb{E}\|\nabla_{z} \mathcal{L}(z_1)-v_1\|^2  \leq \mathcal{J}(z_1) + \frac{\sigma^2}{2 b L},
\end{align}
where the above inequality holds by Assumption \ref{ass:1}.
Since $\eta_t$ is decreasing on $t$, i.e., $\eta_T^{-1} \geq \eta_t^{-1}$ for any $0\leq t\leq T$, we have
 \begin{align}
 \frac{1}{T} \sum_{t=1}^T \mathbb{E} \|\mathcal{P}_{t}\|^2  
 & \leq  \frac{4}{T \epsilon \eta_T} \sum_{t=1}^T\big(\Omega_t - \Omega_{t+1}\big) + \frac{1}{T\eta_T}\sum_{t=1}^T\frac{2\bar{c}\sigma^2}{\epsilon L \hat{c}^2 b}\eta^2_t \nonumber \\
 & \leq \frac{4}{T\epsilon\eta_T} \big( \mathcal{J}(z_1) + \frac{\sigma^2}{ 2b L } - \mathcal{J}(z^*) \big) + \frac{1}{T\eta_T}\sum_{t=1}^T\frac{2\bar{c}\sigma^2}{\epsilon L \hat{c}^2 b}\eta^2_t  \nonumber \\
 & \leq \frac{4 (\mathcal{J}(z_1) - \mathcal{J}(z^*))}{T\epsilon\eta_T} + \frac{2\sigma^2}{bL \epsilon \eta_TT} + \frac{2\bar{c}\sigma^2}{\eta_TT\epsilon L \hat{c}^2 b}\int^T_1\frac{\hat{c}^2}{\bar{c}+t} dt\nonumber \\
 & \leq  \frac{4 (\mathcal{J}(z_1) - \mathcal{J}(z^*))}{T\epsilon\eta_T} + \frac{2\sigma^2}{b L \epsilon \eta_TT} + \frac{2\bar{c}\sigma^2}{\eta_TT \epsilon L b}\ln(\bar{c}+T)\nonumber \\
 & = \bigg( \frac{4 (\mathcal{J}(z_1) - \mathcal{J}(z^*))}{\epsilon \hat{c}} + \frac{2\sigma^2}{b L \epsilon \hat{c}} + \frac{2\bar{c}\sigma^2}{\hat{c} \epsilon L b}\ln(\bar{c}+T) \bigg)\frac{(\bar{c}+T)^{1/2}}{T},
\end{align}
where the second inequality holds by the above inequality \eqref{eq:H6} and the fact that $\mathcal{J}_{T+1} \geq \mathcal{J}^{*}$
Let $G =  \frac{4 (\mathcal{J}(z_1) - \mathcal{J}(z^*))}{\epsilon \hat{c}} + \frac{2\sigma^2}{b L \epsilon \hat{c}} + \frac{2\bar{c}\sigma^2}{\hat{c} \epsilon L b}\ln(\bar{c}+T) $,
we have
\begin{align} \label{eq:66}
 \frac{1}{T} \sum_{t=1}^T \mathbb{E}\big[|\mathcal{P}_{t}\|^2 \big]  \leq \frac{G}{T}(\bar{c}+T)^{1/2}.
\end{align}
According to Jensen's inequality, we have
\begin{align}
 &  \frac{1}{T} \sum_{t=1}^T \mathbb{E}\big[\|\mathcal{P}_{t}\| \big] \leq \bigg( \frac{1}{T} \sum_{t=1}^T \mathbb{E}\big[\|\mathcal{P}_{t}\|^2 \big] \bigg)^{1/2}
 \leq \frac{\sqrt{G}}{T^{1/2}}(\bar{c}+T)^{1/4} \leq \frac{\sqrt{G}\bar{c}^{1/4}}{T^{1/2}} + \frac{\sqrt{G}}{T^{1/4}}
\end{align}
where the last inequality is due to $(a+b)^{1/4} \leq a^{1/4} + b^{1/4}$ for all $a,b>0$. Thus, we have
\begin{align} 
 \frac{1}{T} \sum_{t=1}^T \mathbb{E} \|\mathcal{P}_{t}\|   \leq \frac{\sqrt{G}\bar{c}^{1/4}}{T^{1/2}} + \frac{\sqrt{G}}{T^{1/4}}
\end{align}

\end{proof}

\end{document}